\documentclass[10pt]{article} 
\usepackage[preprint]{tmlr}

\usepackage{amsmath,amsfonts,bm}

\def\eqref#1{equation~\ref{#1}}

\def\1{\bm{1}}

\DeclareMathAlphabet{\mathsfit}{\encodingdefault}{\sfdefault}{m}{sl}
\SetMathAlphabet{\mathsfit}{bold}{\encodingdefault}{\sfdefault}{bx}{n}

\usepackage{hyperref}
\usepackage{url}

\title{Extending Graph Condensation to Multi-Label Datasets: A Benchmark Study}

\author{\name Liangliang Zhang \email zhangl41@rpi.edu \\
      \addr Rensselaer Polytechnic Institute
      \AND
      \name Haoran Bao \email baoh2@rpi.edu \\
      \addr Rensselaer Polytechnic Institute
      \AND
      \name Yao Ma \email may13@rpi.edu\\
      \addr Rensselaer Polytechnic Institute
      }

\usepackage{cleveref}
\usepackage{multirow}
\usepackage{multicol}
\usepackage{makecell}
\usepackage{diagbox}
\usepackage{microtype}
\usepackage{graphicx}
\usepackage{caption}
\usepackage{subcaption}
\usepackage[ruled,vlined,linesnumbered,ruled]{algorithm2e}
\usepackage{amsmath}
\usepackage{float}
\usepackage{tabularx}

\begin{document}
\maketitle
\begin{abstract}
As graph data grows increasingly complicate, training graph neural networks (GNNs) on large-scale datasets presents significant challenges, including computational resource constraints, data redundancy, and transmission inefficiencies. 
While existing graph condensation techniques have shown promise in addressing these issues, they are predominantly designed for single-label datasets, where each node is associated with a single class label. However, many real-world applications, such as social network analysis and bioinformatics, involve multi-label graph datasets, where one node can have various related labels. To deal with this problem, we extends traditional graph condensation approaches to accommodate multi-label datasets by introducing  modifications to synthetic dataset initialization and condensing optimization. Through experiments on eight real-world multi-label graph datasets, we prove the effectiveness of our method. In experiment, the GCond framework, combined with K-Center initialization and binary cross-entropy loss (BCELoss), achieves best performance in general. This benchmark for multi-label graph condensation not only enhances the scalability and efficiency of GNNs for multi-label graph data, but also offering substantial benefits for diverse real-world applications.
\end{abstract}

\section{Introduction}

Graph-structured data are fundamental to many real-world applications, including social networks, academic citation networks, chemical molecules, protein-protein interaction networks, mapping services, and product recommendation systems \cite{battaglia2018relational,wu2020comprehensive,zhou2020graph,wu2022graph}. In these graph structures, nodes represent entities (e.g., users in social networks), while edges represent relationships between these entities (e.g., social connections between users)\cite{zhang2023survey}. 

Graph Neural Networks (GNNs) are a class of deep learning models specifically designed to process and learn from graph-structured data. It utilizes the topology of the graph to capture dependencies and learn meaningful nodes, edges, and graph-level representations. By extending deep learning to non-Euclidean domains, GNNs have achieved state-of-the-arts performance in many graph machine learning applications \cite{zhang2023survey, hamilton2018}. For example, in the recommendation system \cite{Li2019DeepGCNsCG,mao2021ultragcn, wu2022graph,zhang2019star}, drug discoveries \cite{jiang2021could, duvenaud2015convolutional, zhang2021graph}, fraud account detection \cite{dou2020enhancing, wang2021review} and traffic forecasting \cite{jiang2022graph}. 
GNNs combine the node features and structure of graph as the input and output the representations of graph information \cite{hamilton2018, velickovic2017graph, kipf2016semi}. Among all the GNNs techniques, the most broadly followed methods is recursive neighborhood aggregation scheme, which aims to get the representation of each node, respectively \cite{xu2018powerful}. By applying different neighborhood aggregation and graph-level pooling methods, GNNs are capable of handling various of tasks like node classification \cite{xiao2022graph}, link prediction \cite{hasan2011survey}, and graph classification \cite{zhang2018end}.

Even though GNNs\cite{zhang2023survey, zhang2019graph} are widely used for analyzing graphs, training these models on large-scale graphs is still computationally expensive \cite{gao2018large, bojchevski2020scaling}. Furthermore, nowadays, graph data is becoming gigantic with numerous nodes and edges in a single graph \cite{zhang2023survey}. For example, the Twitter user graph has 288M monthly active users as of 3/2015 and an estimated average of 208 followers per user for an estimated total of 60B followers (edges) \cite{ching2015one}. Such industry graphs can be two orders of magnitude larger, like hundreds of billions or up to one trillion edges. In other words, the complexity and large scale of graph data pose significant challenges for training GNNs\cite{GCond}. Despite the scalability challenges, for some specific applications that need re-training multiple times models would encounter more hardships rather than computation cost. 

To address these large-scale datasets challenges, graph condensation methods recently \cite{xu2024survey,gao2024graph} have gained lots of attention. Graph condensation is motivated by condensing a large graph dataset into a smaller informative synthetic graph, thereby enhancing the scalability and efficiency of GNNs \cite{xu2024survey,gao2024graph,Hashemi2024ACS}. The goal is to achieve comparable performance in GNNs using this synthetic small graph, instead of relying on the original large-scale graph \cite{Loukas2018GraphRB,Jin2022CondensingGV}. By doing so, condensed synthetic graph enables GNNs to maintain strong predictive performance while significantly reducing the computational resources required for tasks such as training and inference. By eliminating redundant information in original dataset, it makes the synthetic graph more manageable within the constraints of limited computation resources, thereby providing better support for graph data
mining tasks and applications such as Continual learning \cite{liu2023cat}, Network Architecture Search (NAS)\cite{gao2021graph}, etc. Moreover, take node classification task as an example, the node can be well classified is because GNNs have learned to capture the unique pattern of nodes to distinguish them from other nodes in different classes.

Existing graph condensation works \cite{xu2024survey, gao2024graph} investigate different condensing strategies including gradient distance matching\cite{GCond,SGDD}, trajectory matching\cite{zheng2024structure,zhang2024navigating}, kernel ridge regression\cite{xu2023kernel,wang2024fast}, and distribution matching\cite{GCDM,liu2023cat}. Even though they have achieved great success, the original and synthetic datasets are all single-label ones.  
In many real-world scenarios, the nodes of graph can be associated with multiple labels in lots of situations~\cite{huang2012multi,read2011classifier}. For instance, users are allowed to join multiple groups that respectively represent their diverse interests. Therefore, nodes of social networks \cite{shi2019mlne} are annotated with several tags profiling users' preferences in different domains. Similarly, a paper in a citation network \cite{akujuobi2019collaborative} may be associated with multiple research topics, and in the protein-protein interaction networks \cite{zeng2019graphsaint}, it is used to classify protein functions in biology, which are associated with multiple biological processes, like molecular activities or cellular components. These functions are typically inferred based on how proteins interact with one another, with each interaction providing insight into the role of protein within the cell, such as catalyzing reactions, transporting molecules, or transmitting signals. So in the application of graph datasets, multi-label node classification is a fundamental and practical task in graph data mining\cite{shi2019mlne, shi2020multi, akujuobi2019collaborative, zhao2023multi}. In the multi-label graph scenario, each node not only has content (or features) but also associated with multiple class labels. The goal is to assign one or more labels to each node. However, existing graph condensation techniques are designed to address the single-label graph scalability issue, typically condensing the large graph based on class-wise condensing. In this way, it can be hard for us to get the synthetic small graph from the multi-label graph datasets. 
While directly applying single-label condensation methods to multi-label graphs would result in a loss of critical information, as the class-wise assumption does not hold. This reminds as the open challenge.

To bridge this gap from single-label to multi-label setting, in this work, we adapt available graph condensation techniques to the mainstream multi-label datasets. 
Our approach includes modifications to the synthetic dataset initialization and original datasets condensation matching stages. By evaluating various adaptation settings, we identify three key insights into the optimal combination of initialization and optimization methods. Using these best settings, we further analyze graph condensation models and present three additional observations, highlighting the unique challenges of condensing multi-label graph datasets.

Our contributions are as follows:
\begin{enumerate}
    \item We extend classic SOTA graph condensation methods, including GCond, GCDM, and SGDD, to the multi-label graph dataset scenario by introducing multi-label synthetic dataset initialization and condensing optimization methods.
    \item To find the best adaption strategies, we compare different initialization techniques such as Random sampling, Herding, K-Center and probability synthetic multi-label combine with two different multi-label loss functions -- SoftMarginLoss and BCELoss.
    \item With the best adaption settings, we evaluate the condensation methods with F1-micro and F1-macro scores on eight real-world multi-label datasets: PPI, PPI-large, Yelp, DBLP, PCG, HumanGo, EukaryoteGo, and OGBN-Proteins. Finally, we find the GCond method generally works best with K-Center initialization and BCELoss.
\end{enumerate}

\section{Related Work}
\subsection{Graph Neural Networks}

Graph Neural Networks (GNNs) are a class of deep learning models specifically designed to process and learn from graph-structured data. GNNs have achieved outstanding performance in many graph machine learning applications \cite{zhang2023survey, hamilton2018}. For instance, in the recommendation system on social network \cite{Li2019DeepGCNsCG,mao2021ultragcn, wu2022graph,zhang2019star}, drug discoveries from molecule graphs \cite{jiang2021could, duvenaud2015convolutional, zhang2021graph}, fraud account detection on financial graphs \cite{dou2020enhancing, wang2021review} and traffic forecasting in transportation graph \cite{jiang2022graph}. 

GNNs combine the node features and structure of graph as the input and output the representations of graph \cite{hamilton2018, velickovic2017graph, kipf2016semi}. Among all the GNNs techniques, they broadly follow a recursive neighborhood aggregation scheme to get the representation of each node \cite{xu2018powerful}. Some of the typical GNNs models like the Graph Convolutional Network (GCN) \cite{kipf2016semi} operate by aggregating and transforming information from neighboring nodes layer by layer. Variants like Graph Attenstion Nesworks (GAT) \cite{velickovic2017graph} use attention mechanisms to prioritize certain neighbors, while GraphSAGE \cite{hamilton2017inductive} samples neighborhoods to scale to large graphs. By applying different neighborhood aggregation and graph-level pooling methods, GNNs are capable of handling various of tasks \cite{xiao2022graph,hasan2011survey,zhang2018end} like node classification, link prediction, and graph classification.

\subsection{Multi-Label Classification on Graphs}

Multi-label classification on graphs is a fundamental task in graph machine learning, where nodes in a graph are associated with multiple labels simultaneously. This task is critical in numerous real-world applications, including social networks, bioinformatics, and recommendation systems. For instance, in social networks, users may belong to multiple interest groups, while in bioinformatics, proteins may participate in multiple biological processes such as molecular functions and cellular activities \cite{huang2012multi, read2011classifier, shi2019mlne, akujuobi2019collaborative}. In multi-label classification, the goal is to predict a set of labels for each data point in the dataset. The labels can be thought of as binary variables, where a label is either present (1) or absent (0). Recently, there have been several approaches that use neural networks for multi-label classification. 

Traditional multi-label classification methods relied on adapting single-label classification techniques to multi-label scenarios. Popular strategies include problem transformation methods (e.g., binary relevance and classifier chains) and algorithm adaptation approaches, which extend existing classifiers to handle multiple labels directly \cite{read2011classifier, tsoumakas2008multi}. These approaches use architectures such as recurrent neural networks (RNNs) and convolutional neural networks (CNNs) to learn representations of the input data and make predictions for the labels. However, these approaches were not designed to exploit the graph structure and are therefore suboptimal for graph data.

With the advent of GNNs, researchers began exploring multi-label classification directly on graph-structured data. GNNs excel in capturing the relational and topological information of graphs, making them well-suited for multi-label classification tasks. For example, GraphSAGE \cite{hamilton2017inductive} and GAT \cite{velickovic2017graph} utilize neighborhood aggregation to generate node embeddings, which can be fed into multi-label classifiers. Specialized models like ML-GCN \cite{chen2019multi} integrate graph convolutional layers with multi-label prediction heads to improve performance on multi-label datasets. 

\subsection{Graph Condensation}

Despite the impressive performance of GNN models, training them on large-scale datasets is often challenging due to high computational costs, especially when retraining or dealing with redundant information. To address these issues, graph condensation techniques have been developed. These methods distill a large-scale graph into a smaller yet informative synthetic graph, improving the scalability and efficiency of GNNs \cite{xu2024survey,gao2024graph,Hashemi2024ACS}. The condensation process generates a compact representation that preserves essential topological and feature-based properties from the original graph \cite{Loukas2018GraphRB,Jin2022CondensingGV}. This allows GNNs to achieve comparable performance using the condensed graph while significantly reducing the computational resources needed for training and inference. 

Existing graph condensation works \cite{xu2024survey, gao2024graph} investigate different condensing strategies including gradient distance matching\cite{GCond,SGDD}, trajectory matching\cite{zheng2024structure,zhang2024navigating}, kernel ridge regression\cite{xu2023kernel,wang2024fast}, and distribution matching\cite{GCDM,liu2023cat}. Condensed graphs facilitate more efficient model execution without sacrificing the quality of learning, making this technique a promising method for real-world GNN deployments. By eliminating redundant information in original dataset, graph condensation makes the synthetic graph more manageable within the constraints of limited computation resources, thereby providing better support for graph data
mining tasks and applications such as Continual learning \cite{liu2023cat} and Network Architecture Search (NAS)\cite{gao2021graph}, etc. Moreover, take node classification as an example, the reason a node can be well classified is that GNNs have learned to capture the unique pattern of nodes to distinguish them from other nodes in different classes.
\section{Problem Formulation}
\begin{figure*}[h]
  \centering
  \includegraphics[width=\linewidth]{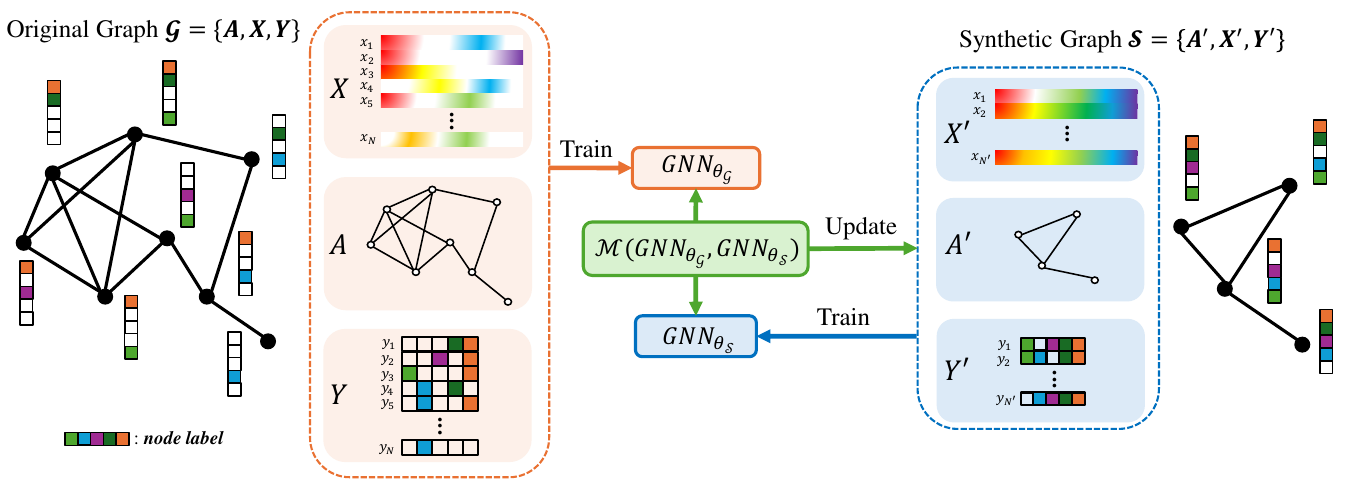}
  \caption{Workflow for multi-label graph condensation. It shows the process of condensing a large multi-label graph $\mathcal{G}=\{A, X, Y\}$ into a smaller synthetic graph $\mathcal{S}=\{A', X', Y'\}$, where $Y$ and $Y'$ represent the multi-label matrix. Various matching strategies, denoted by $\mathcal{M}(\cdot)$, are employed to ensure that key information in the original graph is captured. Our goal is to use the synthetic graph $\mathcal{S}$ to train a GNN that achieves comparable performance to the one trained on the original graph $\mathcal{G}$, thus reducing graph size while retaining performance.}
  \label{fig:workflow}
\end{figure*}
Consider a graph $\mathcal{G}=\{A, X, Y\}$, where $N$ is the number of nodes, $A\in\mathbb{R}^{N \times N}$ is the corresponding adjacency matrix,  $X = \{x_1, x_2, \cdots, x_N\} \in \mathbb{R}^{N \times d}$ is the $d$-dimensional node feature matrix. In multi-label scenario, each node is associated with multiple labels revealing characteristics or semantics of the node. In this case, the label matrix can be represented as $Y\in \{y_1, y_2, \cdots, y_N \} \in \{0,1\}^{N \times K}$, where $K$ is the number of classes. Following the definition of graph condensation, our goal is to learn a smaller synthetic graph denoted as $\mathcal{S}$, which can contain the ability to train downstream tasks' GNNs and achieve competitive performance compared with the original large-scale complex graph $\mathcal{G}$. Similarly, the synthetic graph can be denoted as $\mathcal{S}=\{A', X', Y'\}$ along with $A' \in\mathbb{R}^{N' \times N'}$, and $X' \in \mathbb{R}^{N' \times d}$, $Y' \in \{0, 1\}^{N' \times K}$. During this stage, our expectation is that the scale of the synthetic graph can be much smaller compared with the original graph, denotes as $N'\ll N$.

In this work, we adapt the graph condensation techniques to a multi-label setting, trying to get the synthetic multi-label graph. First, we introduce the general framework of graph condensation in Figure \ref{fig:workflow}. Generally, graph condensation methods are using different matching strategies to optimize the synthetic graph $\mathcal{S}$ similar to the original graph $\mathcal{G}$. Despite these various strategies, the key point is to keep the comparable performance of GNNs while training on small graph. Therefore, our objective can be formulated as follows \cite{GCond}:
\begin{equation}
    \mathcal{S^*} = \arg \min_{\mathcal{S}} \mathcal{M}(\theta_{\mathcal{S}}^*, \theta_{\mathcal{G}}^*) \quad \text{s.t.} \quad \theta^* = \arg \min_{\theta} \mathcal{L}({GNN}_\theta),
\end{equation}
where ${GNN}_\theta$ denotes a GNN parameterized with $\theta$. Specifically, $\theta_{\mathcal{S}}$ and $\theta_{\mathcal{G}}$ are the parameters trained on graph $\mathcal{S}$ and $\mathcal{G}$, respectively. $\mathcal{M}(\cdot)$ is the matching strategy used to match the model $\theta_{\mathcal{S}}^*$ trained synthetic graph $\mathcal{S}$ to the one $\theta_{\mathcal{G}}^*$ trained on the original graph $\mathcal{G}$. $\mathcal{L}(\cdot)$ is the loss function used to measure the difference between model predictions and ground truth. For most graph condensation methods, the loss function is cross-entropy loss.

As the specific initialization of model parameters $\theta$ can lead to the overfitting problem \cite{GCond}, one common solution \cite{wang2018dataset} is to generate the synthetic data following a distribution of random initializations of $P_{\theta_0}$. Furthermore, the reformulated optimization problem as follows:
\begin{equation}
    \begin{aligned}
    \mathcal{S^*} = \arg \min_{\mathcal{S}} \mathbb{E}_{\theta_0 \sim  P_{\theta_0}} [\mathcal{M}(\theta_{\mathcal{S}}^*, \theta_{\mathcal{G}}^*) \quad \\ \text{s.t.} \quad \theta^* = \arg \min_{\theta(\theta_0)} \mathcal{L}({GNN}_\theta)].
    \end{aligned}
\end{equation}
where $\theta(\theta_0)$ indicates $\theta$ is a function acting on $\theta_0$.

\section{Graph Condensation Methods}\label{sec:single-label-condensation}

Building on the global understanding of existing single-label graph condensation methods, we observe that while different graph condensation methods differ primarily in their matching strategies, the overall workflow remains largely consistent. Therefore, before adapting these methods to the multi-label scenario, we will illustrate the different matching strategies of condensation methods first.

\subsection{GCond: Gradient Matching}
By computing the gradient of $\mathcal{L}$ w.r.t. $\mathcal{S}$ and optimize $\mathcal{S}$ via gradient descent. For each training step, the $opt_\theta$ is the update rule, consider the one-step gradient descent for model parameters:
\begin{equation}
    \begin{aligned}
        \theta^{\mathcal{S}}_{t+1} \gets \theta^{\mathcal{S}}_{t} - \eta \nabla_{\theta}\mathcal{L}({GNN}_{\theta^{\mathcal{S}}_t}(A', X'), Y') \\ \theta^{\mathcal{G}}_{t+1} \gets \theta^{\mathcal{G}}_{t} - \eta \nabla_{\theta}\mathcal{L}({GNN}_{\theta^{\mathcal{G}}_t}(A, X), Y)
    \end{aligned}
\end{equation}
where $\theta^{\mathcal{S}}_{t}$ and $\theta^{\mathcal{G}}_{t}$ denote the model parameters trained on $\mathcal{S}$ and $\mathcal{G}$ at step $t$. $\eta$ is the learning rate for the gradient descent.

By matching the training gradient trajectory, define distance function $D(\cdot, \cdot)$, define the gradient matching strategy $\mathcal{M}_{GCond}$ as:
\begin{equation}
    \begin{aligned}
        \mathcal{M}_{GCond} = D(\nabla_{\theta}\mathcal{L}({GNN}_{\theta^{\mathcal{S}}_t}(A', X'), Y'), \\
        \nabla_{\theta}\mathcal{L}({GNN}_{\theta^{\mathcal{G}}_t}(A, X), Y))
    \end{aligned}
\end{equation}

Assume $T$ is the number of steps of the whole training, the final optimized objective is:
\begin{equation}
    \begin{aligned}
        \mathcal{S^*} = \arg \min_{\mathcal{S}} \mathbb{E}_{\theta_0 \sim  P_{\theta_0}} [\sum_{t=0}^{T-1}D(\nabla_{\theta}\mathcal{L}({GNN}_{\theta^{\mathcal{S}}_t}(A', X'), Y'), \\
        \nabla_{\theta}\mathcal{L}({GNN}_{\theta^{\mathcal{G}}_t}(A, X), Y))].
    \end{aligned}
\end{equation}

Direct joint learning can be highly challenging because the three variables $A', X'$, and $Y'$ in synthetic graph are interdependent. One straightforward way is to treat $A'$ and $X'$ as free parameters. By modeling the synthetic graph structure as a function of the condensed node features, synthetic graph structure $A'$ can be learned by:
\begin{equation}
    A' = g_\phi(X') \label{eq:structure}
\end{equation}
where $\phi$ is the parameter of a multi-layer neural network. So the rewrite objective is:
\begin{equation}
    \begin{aligned}
        \mathcal{S^*} = \arg \min_{X',\phi} \mathbb{E}_{\theta_0 \sim  P_{\theta_0}} [\sum_{t=0}^{T-1}D(\nabla_{\theta}\mathcal{L}({GNN}_{\theta^{\mathcal{S}}_t}(g_\phi(X'), X'), Y'), \\
        \nabla_{\theta}\mathcal{L}({GNN}_{\theta^{\mathcal{G}}_t}(A, X), Y))]
    \end{aligned}
\end{equation}
\subsection{SGDD: Structure-broadcasting}
Instead of treating $A'$ and $X'$ in the synthetic graph $\mathcal{S}$ as free parameters, recent work \cite{SGDD} points out that this way ignores the original structure $A$ in $\mathcal{G}$. To address this problem, they broadcast $A$ as supervision for the generation of $A'$. By introducing graphon \cite{gao2019graphon,ruiz2020graphon,xia2023implicit} to matching the different shapes between $A$ and $A'$, SGDD use random noise $\mathcal{Z}(N')\in \mathbb{R}^{N' \times N'}$ as input coordinates. Through the generative model to synthesize an adjacency matrix $A'$ with $N'$ nodes, the process can be formulated as:
\begin{equation}
    A' = GEN(\mathcal{Z}(N'); \Phi)
\end{equation}
where $GEN(\cdot)$ is the generative model with parameter $\Phi$. Then the structure optimization is:
\begin{equation}
    \mathcal{M}_{structure} = Dis(A, GEN(\mathcal{Z}(N'); \Phi)),
\end{equation}
where $A$ is supervision and $Dis(\cdot)$ is a metric that measure the difference between $A$ and $A'$. By adding the corresponding nodes information from $X'$ and $Y'$ to the inherent relation learning \cite{pfeiffer2014attributed,shalizi2011homophily}, the revised generation model is $GEN(\mathcal{Z}(N')\oplus X' \oplus Y'; \Phi)$, $\oplus $ denotes the concatenate operation.
In general, SGDD concurrently optimizes the parameters $X'$ and $A'$. While the refinement of $X'$ is achieved through a gradient matching strategy, whereas the $A'$ is enhanced using the Laplacian
energy distribution(LED) matching technique \cite{tang2022rethinking,das2016distribution,gutman2006laplacian}. During each step, the other component is frozen to ensure effective refinement, and the overall training loss function can be summarized as:
\begin{equation}
    \mathcal{M}_{SGDD} = \mathcal{M}_{GCond} + \alpha \mathcal{M}_{structure}+\beta{||A||}_2,
    \label{eq:SGDD}
\end{equation}
where ${||A||}_2$ is proposed as a sparsity regularization term, $\alpha$ and $\beta$ are trade-off parameters.

\subsection{GCDM: Distribution Matching}
Inspired by the distribution matching strategy in image dataset condensation \cite{zhao2023dataset}, GCDM\cite{GCDM} adapt the receptive field distribution matching method in GNN. 
In the graph dataset, with the reproducinf kernal Hilbert space $\mathcal{H}$ defined by GNNs and with in each class $c$, GCDM optmize the maximum mean discrepancy (MMD). Let $\Phi_\theta$ be an $L-$layer GNN model parameterized by $\theta$, the loss of GCDM is:
\begin{equation}
    \mathcal{M}_{GCDM} = \min_{\phi,X'}\sum_{c=0}^{C-1}r_c \cdot \max_{\theta_c}\left \| \frac{1}{V_c} \sum_{i \in V_c} \text{emb}_{i}^{c} -  \frac{1}{V_c'} \sum_{j \in V_c'} \text{emb}_{j}^{c}\right \|^2_2 ,
\end{equation}
where the node embeddings are:
\begin{equation}
    \{\text{emb}_{i}^{c}\}_{i=0}^{N-1} \gets \Phi_{\theta_c}(A, X)
\end{equation}
\begin{equation}
    \{\text{emb}_{j}^{c}\}_{j=0}^{N'-1} \gets \Phi_{\theta_c}(A' = g_\phi(X'), X')
\end{equation}
$V_c$ and $V_c'$ represent the node sets belong to the class $c$ of original graph $\mathcal{G}$ and synthetic graph $\mathcal{S}$, respectively. $r_c$ denotes the condensed class ratio for each class of the synthetic graph.

\section{Datasets and Experiment Settings}

In this section, we first present the multi-label graph datasets used in our benchmark investigation, accompanied by the detailed introduction of label information in real-world application. Next, we detail the evaluation metrics used to assess the effectiveness of the adaptation methods, providing a robust framework for performance comparison.

\noindent{\bf Datasets.}
 Specifically, we employ eight real-world datasets including PPI, PPI-large, Yelp, DBLP, OGBN-Proteins, PCG, HumanGo, EukaryoteGo. The details of those datasets are introduced as follows:

\begin{itemize}

\item PPI \cite{zeng2019graphsaint} is the Protein-Protein Interaction network. It classifies protein functions based on the interactions of human tissue proteins. Positional gene sets are used, motif gene sets and immunological signatures as features and gene ontology sets as labels (121 in total), collected from the Molecular Signatures Database \cite{liberzon2015molecular}. And PPI-large is the larger version of PPI dataset. 
\item Yelp \cite{zeng2019graphsaint} is processed from a public open dataset from \href{https://www.yelp.com/dataset}{yelp.com}. The nodes represent active users and the edges are the relationship between the users. The multi-label of each node represents the types of business like Coffee \& Tea, Flowers \& Gifts, Tours, and so on. Node features contain the information of all the reviews and rates by users, generated by pre-trained Word2Vec model \cite{church2017word2vec}.

\item DBLP \cite{akujuobi2019collaborative} is the citation network extracted from DBLP. The nodes represent authors and edges are the co-authorship between the authors. Multi-label here indicates the four research areas like database, data mining, information and retrieval and artificial intelligence.

\item OGBN-Proteins from Open Graph Benchmark \cite{hu2020ogb} is an undirected, weighted, and typed (according to species) graph. Nodes represent proteins, and edges indicate different types of biologically meaningful associations between proteins, e.g., physical interactions, co-expression or homology. The task is to predict the presence of protein functions in a multi-label binary classification setup, where there are 112 kinds of protein functions denote as labels to predict in total. Here the protein functions refer to the specific biological activities or roles that a protein performs within a cell or organism. These functions are typically described using Gene Ontology (GO) database \cite{huntley2015goa} or other biological annotation systems and can fall into different categories.

\item  PCG \cite{zhao2023multi} is Protein-Phenotype graph dataset, which focused on predicting phenotypes associated with proteins. Phenotypes refer to observable traits or characteristics of diseases, and identifying these associations can be valuable for clinical diagnostics or discovering potential drug targets. The dataset is constructed using protein-phenotype associations from the DisGeNET database \cite{pinero2020disgenet}, where each protein is linked to one or more phenotypes. The phenotypes are grouped into disease categories based on the MESH ontology \cite{bhattacharya2011mesh}, and labels with fewer than 100 associated proteins are removed. The PCG includes 3,233 proteins as nodes and 37,351 protein-protein interactions as edges. The task is to predict multiple phenotype labels for each protein based on its interactions and features.
\item HumanGo \cite{chou2007euk,liberzon2015molecular} contains 3,106 proteins, each potentially associated with one or more of 14 subcellular locations as multi-label. The proteins are represented as nodes in a graph, with features generated from their sequences using pre-trained model \cite{yang2020prediction}. Protein-protein interactions form the edges of the graph, with each edge representing the confidence of different interaction types. Different location associations come from the GO database \cite{huntley2015goa}, which assigns standardized terms to proteins based on their roles in biological processes (e.g., cell division, metabolism), molecular functions (e.g., enzyme activity, receptor binding), and cellular components (e.g., nucleus, membrane).
\item EukaryoteGo \cite{chou2007euk,liberzon2015molecular} follows a similar structure but focuses on eukaryote proteins. It includes 7,766 proteins and their interactions, with proteins being assigned to one or more of 22 subcellular locations. Like HumanGo, it uses protein sequences for node features and protein-protein interactions for graph structure. 

\end{itemize}

We follow the predefined data splits from \cite{zeng2019graphsaint, zhao2023dataset, hu2020ogb, chou2007euk}. \ref{tab:dataset} presents an overview of the datasets' characteristics.

\begin{table}[tbhp]
\footnotesize
\caption{Dataset Statistics}\label{tab:dataset}
\begin{center}
\resizebox{\columnwidth}{!}{ 
\begin{tabular}{c c c c c c c } \hline
\bf Dataset & \bf \#Nodes & \bf \#Edges & \bf \#Avg.Edges & \bf \#Features & \bf \#Labels & \bf Train/Val/Test \\ \hline
\bf PPI	& 14,755 & 225,270 &15.27 & 50 & 121 & 0.66/0.12/0.22 \\
\bf PPI-large & 56,944  & 818,716 &14.38 & 50 & 121 & 0.79/0.11/0.10 \\
\bf Yelp & 716,847 & 6,977,410 &9.73 & 300 & 100 & 0.75/0.10/0.15 \\
\bf DBLP & 28,702 & 68,335 &2.38 & 300 & 4 & 0.60/0.20/0.20 \\
\bf OGBN-Proteins & 132,000	& 39,000,000 &295.45 & 8 &  112	& 0.66/0.16/0.18 \\
\bf PCG & 3,233	& 37,351 &11.55 & 32 & 15 & 0.60/0.20/0.20 \\
\bf HumanGo	& 3,106	& 18,496 &5.96 & 32 & 14 & 0.60/0.00/0.40 \\
\bf EukaryoteGo	& 7,766	& 13,818 &1.78 & 32 & 22 & 0.60/0.00/0.40 \\
\hline
\end{tabular}
}
\end{center}
\end{table}
\noindent{\bf Metrics.}
To evaluate the performance of adapted graph condensation methods, we utilize two widely recognized metrics in multi-label classification: F1-micro and F1-macro.

\begin{itemize}
    \item F1-micro score aggregates the contributions of all classes to compute the average F1 score. It does this by first calculating the total true positives, false positives, and false negatives across all labels, and then using these aggregated values to compute a global precision and recall. The F1-micro score is particularly useful when dealing with imbalanced datasets, as it gives equal weight to each instance rather than each class. It is defined as:
    \begin{equation}
    \text{F1-micro} = \frac{2 \times \text{Precision} \times \text{Recall}}{\text{Precision} + \text{Recall}}
    \end{equation}
    
    where:
    \begin{align*}
    \text{Precision} & = \frac{\text{TP}}{\text{TP} + \text{FP}} \\
    \text{Recall} & = \frac{\text{TP}}{\text{TP} + \text{FN}}
    \end{align*}

    \item F1-macro score computes the F1 score for each class independently and then averages these scores. This metric treats all classes equally, regardless of their frequency in the dataset. It is particularly useful for understanding the model's performance across all classes, especially in scenarios where some classes may be underrepresented. The F1-macro score is defined as:

    \begin{equation}
    \text{F1-macro} = \frac{1}{C} \sum_{i=1}^{C} \text{F1}_{i}
    \end{equation}
    
    where \( C \) is the number of classes and \(\text{F1}_{i}\) is the F1 score for the \(i^{th}\) class, calculated as:
    
    \begin{equation}
    \text{F1}_{i} = \frac{2 \times \text{Precision}_{i} \times \text{Recall}_{i}}{\text{Precision}_{i} + \text{Recall}_{i}}
    \end{equation}
\end{itemize}

After introduce the single-label graph condensation methods, we will focus on adjusting key components to fit the multi-label task while preserving the core matching mechanisms. This ensures that our framework can be extended to other existing graph condensation techniques, allowing for flexible and efficient adaptation across different methods.
This benchmark presents a thorough evaluation of the various settings and methods used for graph condensation in the multi-label scenario. We structure the benchmark results into three key parts: (1) identifying the best initialization and loss function settings, (2) comparing the full results of all graph condensation methods across multiple datasets, and (3) analyzing the results through multi-label class distribution and label correlation.

\section{Adapt Existing Graph Condensation Methods for Multi-label Datasets}
\label{sec:1}
\begin{table*}[t]
    \centering
    \resizebox{\textwidth}{!}{
    \begin{tabular}{c c c c c c c c c c}
    \hline
    \multirow{2}{*}{\bf Datasets} & \multirow{2}{*}{\bf C-rate} & \multicolumn{2}{c}{\bf Random} & \multicolumn{2}{c}{\bf Herding} & \multicolumn{2}{c}{\bf K-Center} & \multirow{2}{*}{\bf Probability} & \multirow{2}{*}{\bf Whole Dataset}\\
    \cline{3-8}
    & & \bf Subgraph & \bf Nodes & \bf Subgraph & \bf Nodes & \bf Subgraph & \bf Nodes & & \\
     \hline
    \bf PPI& \bf 1.00\%&    47.05&	\textbf{40.49}&	40.93&	39.47& \textbf{47.77}&	36.60&	40.04&	51.26\\
    \bf Yelp&\bf 0.02\%&	34.21&	30.35&	32.77&	30.96& \textbf{35.60}&	\textbf{31.46}&	20.95&	37.97\\
    \bf DBLP&\bf 0.80\%&	\textbf{63.56}&	\textbf{56.03}&	41.54&	41.02&	60.36&	55.62&	41.35&	87.55\\
    \bf OGBN-Proteins&\bf 0.1\%&	14.12&	15.44&	11.62&	15.36&\textbf{29.95}&	\textbf{24.81}&	15.53&	18.86\\
    \bf PCG&\bf 4\%&	13.94&	19.01&	\textbf{27.76}&	16.35&	25.13&	\textbf{22.18}&	21.69&	42.26\\
    \hline
    \end{tabular}
    }
    \footnotesize
    \caption{F1-Micro Score (\%) of Coreset Method with Different Initialization Strategies}\label{tab:init}
\end{table*}

\begin{table*}
    \centering
    \footnotesize
    \resizebox{\textwidth}{!}{
    \begin{tabular}{c c c c c c c c c c c}
    \hline
    \multirow{2}{*}{\bf Datasets} & \multirow{2}{*}{\bf C-rate} & \multicolumn{2}{c}{\bf Random} & \multicolumn{2}{c}{\bf Herding} & \multicolumn{2}{c}{\bf K-Center} & \multicolumn{2}{c}{\bf Probability} & \multirow{2}{*}{\bf Whole Dataset}\\
    \cline{3-10}
    & & \bf Without $A'$ & \bf With $A'$ & \bf Without $A'$ & \bf With $A'$ & \bf Without $A'$ &\bf With $A'$ &\bf Without $A'$ &\bf With $A'$ & \\
    \hline
    \textbf{PPI} & \textbf{1.00\%} & \textbf{48.55 }&  \textbf{50.17 }& 46.71 & 47.80 & 35.99 & 48.38 & 44.62 & 43.31 & 51.26 \\
    \textbf{Yelp} & \textbf{0.02\%} & 30.16 & 29.53 & 30.48 & 32.23 & \textbf{34.50} & \textbf{34.90} & 21.33 & 22.10 & 37.97 \\
    \textbf{DBLP} & \textbf{1\%} & 52.10 & 57.48 & 45.95 & 47.12 & \textbf{52.56} & \textbf{60.77} & 48.24 & 48.56 & 87.55 \\
    \textbf{OGBN-Proteins} & \textbf{0.10\%} & 22.03 & 28.69 & 20.54 & 25.19 & \textbf{28.09} & \textbf{29.04} & 22.10 & 27.01 & 30.59 \\
    \textbf{PCG} & \textbf{4\%} & 20.50 & 23.79 & \textbf{22.14} & 25.38 & 19.88 & 22.36 & 22.07 & \textbf{27.21} & 42.26 \\
    \hline
    \end{tabular}
    }
    \captionsetup{justification=centering}
    \caption{F1-Micro Score (\%) of GCond Method with Random/Herding/K-Center/Probability Distribution Initialization with/without Learning from Structure for SoftMarginLoss.}
    \label{tab:loss1}
\end{table*}

\begin{table*}
    \centering
    \footnotesize
    \resizebox{\textwidth}{!}{
    \begin{tabular}{c c c c c c c c c c c}
    \hline
    \multirow{2}{*}{\bf Datasets} & \multirow{2}{*}{\bf C-rate} & \multicolumn{2}{c}{\bf Random} & \multicolumn{2}{c}{\bf Herding} & \multicolumn{2}{c}{\bf K-Center} & \multicolumn{2}{c}{\bf Probability} & \multirow{2}{*}{\bf Whole Dataset}\\
    \cline{3-10}
    & & \bf Without $A'$ & \bf With $A'$ & \bf Without $A'$ & \bf With $A'$ & \bf Without $A'$ &\bf With $A'$ &\bf Without $A'$ &\bf With $A'$ & \\
    \hline
    \textbf{PPI} & \textbf{1.00\%} & \textbf{49.95} & \textbf{51.35} & 47.47 & 48.23 & 36.79 & 49.27 & 45.65 & 43.91 & 51.26 \\
    \textbf{Yelp} & \textbf{0.02\%} & 29.24 & 30.76 & 31.76 & 32.41 & \textbf{32.36 }&\textbf{33.94} & 22.96 & 25.62 & 37.97 \\
    \textbf{DBLP} & \textbf{1\%} & 53.41 & 59.40 & 46.87 & 48.50 & \textbf{54.21} & \textbf{70.74} & 49.73 & 51.33 & 87.55 \\
    \textbf{OGBN-Proteins} & \textbf{0.10\%} & 24.65 & 28.88 & 21.45 & 26.21 & \textbf{28.15} & \textbf{29.40} & 21.73 & 27.11 & 30.59 \\
    \textbf{PCG} & \textbf{4\%} & 22.26 & 25.45 & \textbf{24.66} & 26.76 & 20.17 & 25.58 & 23.47 &\bf 28.28 & 42.26 \\
    \hline
    \end{tabular}
    }
    \caption{F1-Micro Score (\%) of GCond Method with Random/Herding/K-Center/Probability Initialization with/without Learning from Structure for BCELoss.}
    \label{tab:BCEloss}
\end{table*}

After the introduction of different graph condensation methods based on the matching strategies in Section~\ref{sec:single-label-condensation} including GCond, SGDD and GCDM, we  apply the above methods in multi-label graph datasets in this section. However, these methods could not directly be applied to condense multi-label graphs majorly due to two reasons: (1) The synthetic graph initialization is designed for single-class augmentation with pre-defined condensed labels. (2) The condensation methods depend on GNNs to match the original graph $\mathcal{G}$ and synthetic graph $\mathcal{S}$, the classification objectives are designed for single-label node classification. Therefore, in this section, we would like to adapt the existing graph condensation methods by investigating different adaptation strategies to address these two issues. In particular, we first describe the existing issues and the corresponding resolvents in Section \ref{subsec:adapation}. Then, we conduct experiments to investigate which strategies work best for the multi-label graph condensation tasks in Section \ref{subsec:settings}. 

\subsection{Multi-label Adaptation}
\label{subsec:adapation}
During condensation stage,
among all the condensation methods, class-wise batch sampling is widely used for synthetic graph optimization. For instance, in GCond, they initialize the synthetic graph labels $Y'$ by predefine the labels for each class with condensation rate $r$. While optimization, to further reduce the memory usage and simplify the process, they calculate the gradient matching loss for nodes from different classes separately. As matching the gradients from a single class is easier than from all the classes, most of condensation researches follow this class-wise sampling method. However, in contrast to a single-labeled graph, a multi-labeled graph has multiple class labels assigned to each node, so the class-wise initialization and optimization process need to be revised to the multi-label scenario. Furthermore, when adapting single-label graph condensation techniques to multi-label node classification, the choice of the loss function is crucial to ensuring that the model effectively learns the complex relationships inherent in multi-label tasks. In single-label graph condensation, methods typically optimize for cross-entropy loss, which is suitable for tasks where each node belongs to a single class. However, in multi-label classification, nodes can belong to multiple classes simultaneously. 

Therefore, to better determine the best setting for multi-label graph condensation we investigate different initialization and optimization methods to evaluate their impact on preserving multi-label relationships and classification accuracy. 

Our objective is to assess how these initialization methods and loss functions influence the performance of the condensed graphs in multi-label classification tasks. To sum up, we focus on two key areas of adaptation: \textbf{initialization} and \textbf{optimization} of the synthetic graph. These modifications allow us to capture and preserve the essential multi-label features of the original graph.

\noindent{\bf \underline{Issue 1:} Initialization of Synthetic Multi-Label Graph.}
In single-label settings, the condensed label $Y_{single}'$ is predefined using ground-truth class distribution and retrieves $X'$ randomly from each class.

As most of the graph condensation methods adapt Eq.\ref{eq:structure} to learn synthetic graph structure $A'$ from node features $X'$, one straightforward method for multi-label adaptation is sampling the nodes from original datasets. Inspired by the heuristic sampling algorithms of baseline Coreset methods used in GCond, the sampled subgraph by different strategies can be used for the synthetic multi-label initialization to get a better performance. 

As some of the coreset methods gain comparable performance as baseline \cite{welling2009herding, farahani2009facility, sener2017active}, we apply those heuristic algorithms to extract $N'$ nodes from the full set of nodes to obtain subgraphs $S'$ with corresponding condensation rates as the initialization methods. 

\begin{itemize}
    \item Random selection is sampling the nodes randomly, so the probability of selecting node $i$ is $P(i)=\frac{1}{N}$.
    \item Herding \cite{welling2009herding} method is more structured compared with random selection strategy. It is based on iterative minimization of the distance between the mean of the selected nodes' feature vectors and the mean of the entire graph feature vectors. Let $\mu_{\mathcal{G}}$ present the mean feature vector of the original graph:
    \begin{equation}
        \mu_{\mathcal{G}} = \frac{1}{N}\sum_{i=0}^{N-1}x_i \quad\text{and} \quad \mu_{\mathcal{S}} = \frac{1}{N'}\sum_{j=0}^{N'-1}x'_j
    \end{equation}
    \item K-Center \cite{farahani2009facility,sener2017active} method aims to maximize the coverage of the original graph feature space by minimizing the maximum distance between unselected nodes and their closest selected node. Let $\mathcal{S}=\{s_1, s_2, \cdots, s_{N'}\}$ be the selected node set. The goal is to select nodes such that for every unselected node $i$, the distance to the nearest selected node $s_j \in \mathcal{S}$ is minimized:
    \begin{equation}
        \min_{s_1,\cdots, s_{N'}}\max_{i\in\{1,\cdots, N\}}\min_{j\in \{1,\cdots, N'\}} \left \| x_i - x_{s_j} \right \|_2
    \end{equation}
    The selection process would start from an arbitrary node as the first center $s_1$ and then iteratively select the next node $s_t$ which is farthest away from the selected nodes. The process continues until $N'$ nodes are selected.  
    \item In the multi-label scenario, each node has multiple labels, and it is also important to ensure that the selected subgraph maintains the label distribution of the original graph. Here, we introduce probabilistic synthetic multi-label method, for each class $k$, define the label distribution as the fraction of nodes in the graph that belongs to class $k$:
    \begin{equation}
        p_k = \frac{1}{N}\sum_{i=1}^{N}Y_{i, k}
    \end{equation}
    The goal is to ensure that the synthetic graph $\mathcal{S}$ maintains a similar label distribution. So for each synthetic label $y_j'$ in $ Y'\in{0,1}^{N' \times K}$, the label of class $k$ has a $p_k$ probability to be 1. Then we choose cosine similarity matching for the node features based on the real labels from $\mathcal{G}$ and probability synthetic labels.
    
\end{itemize} 
Above all, in multi-label settings, we adapt subgraph initialization and probabilistic synthetic multi-label methods for multi-label graph condensation. Besides, in graph condensation, the structure $A'$ of synthetic graph is optimized from nodes features $X'$. Based on the researches \cite{ding2022data,zhu2021survey, GCond}, this "Graphless" learning process can be highly useful for downstream data analysis, which also shows competitive performance. Therefore, by comparing different initialization methods with or without learning the graph structure, we get the best settings for the adaptation.

\noindent{\bf \underline{Issue 2:} Optimization about Multi-Label Classification Loss.}

In single-label settings, while the condensed graph is matched in various ways, the fixed label distribution can be used directly for synthetic graph initialization. With predefined synthetic label $Y'_{single}\in \{0, \cdots, C-1\}^{N'}$, the synthetic node feature is randomly retrieved from the particular class. As calculate the matching loss for nodes from different classes separately,is easier than that from all classes. In other words, this single-label class-wise sample method will directly sample each class to optimize. Specifically, for a given class $c$, it will sample a batch of nodes of class $c$ from original graph $\mathcal{G}$ with a portion of neighbors, denotes as $(A_c, X_c, Y_c) \sim \mathcal{G}$. For synthetic graph $S$, only sample the specific nodes of class $c$ without the neighbors. By doing so, the synthetic graph will use all the other nodes for aggregation stages during condensation, similarly, this process can be wrote as $(A_c', X_c', Y_c') \sim \mathcal{S}$.
Further in terms of details, for each class $c$, there is a subset of nodes $N_c$ that belong to this class. The loss is computed class-wise, for example, the cross-entropy loss for single node $j$ with ground truth label $y_j$ and predicted logit $z_{j,c}$ for class $c$ is:
\begin{equation}
    \ell_{CE}(z_j, y_j) = - \log(\frac{e^{z_{j,y_j}}}{\sum_{c=0}^{C-1}e^{z_{j,c}}})
\end{equation}
The single-label final loss would be formulated as:
\begin{equation}
    \mathcal{L}_{single-label} = \sum_{c=0}^{C-1} \sum_{j\in N_c}\ell_{CE}(z_j, y_j)
\end{equation}

However, in multi-label classification $Y'\in \{y_1', \cdots, y_N' \} \in \{0,1\}^{N' \times K}$, each node can associate with multiple classes simultaneously. Typically, the multi-label classification task is modeled as $K$ independent binary classification tasks, for each task $k$, the model predicts a logit $z_{j,k}$ and ground-truth label $y_{j,k}$ indicates whether node $j$ belongs to class $k$. 
In multi-label classification,
SoftMarginLoss \cite{cao2019learning} incorporates margin-based optimization, which is beneficial when dealing with imbalanced datasets and overlapping classes. In multi-label settings, certain labels may frequently co-occur or exhibit correlations. SoftMarginLoss introduces a margin between positive and negative label predictions, encouraging the model to differentiate between these labels more confidently.
\begin{equation}
    \mathcal{L}_{Softmargin} (z_{j,k} , y_{j,k})= log(1+e^{- z_{j,k}\cdot y_{j,k}} )
\end{equation}
This formulation is particularly useful in ensuring that the model not only predicts correct labels but does so with confidence, reducing the chances of ambiguous predictions near the decision boundary. 
On the other hand, Binary Cross-Entropy Loss \cite{durand2019learning} is a widely-used objective function for multi-label classification, as it independently evaluates each label for each node in the graph. Then the task-wise loss for each binary classification task is computed using Binary Cross-Entropy (BCE) \cite{durand2019learning} loss:
\begin{equation}
    \mathcal{L}_\text{BCE}(z_{j,k}, y_{j,k}) = -(y_{j,k} \log(\sigma(z_{j,k})) + (1-y_{j,k})\log(1 - \sigma(z_{j,k}))),
\end{equation}
where $\sigma(z_{j,k})=\frac{1}{1+e^{-z_{j,k}}}$ is the sigmoid function. 

\subsection{Experiments and Results Analysis}\label{subsec:settings}

In this section, after the investigation of different initialization and optimization strategies. We combine them together to find out the best settings for multi-label graph condensation adaptation. Additional results in terms of other metrics are presented in Appendix \ref{appendix}.

In this section, we conduct experiments on GCond to investigate the effectiveness of different initialization and optimization strategies on the representative datasets and corresponding condensation rates. We first introduce the experiment settings with specific methods and datasets. In graph condensation, the condensation ratio (often denoted as C-rate) is a measure of how much the original graph has been reduced in size for training purposes. This ratio represents the percentage or fraction of nodes retained from the full dataset in the condensed graph. For example, if the C-rate is set to 1\%, it means that the scale of synthetic graph $\mathcal{S}$ is only 1\% of original graph $\mathcal{G}$. Then we statistically analyzed the experimental results and presented the comparison of different settings for multi-label graph condensation.

\noindent{\bf Settings.} 
For dataset selection, we pick the most representative datasets to simplify the choosing stage, for instance, PPI-large is the larger version of PPI dataset, so we only test the settings in PPI. With the representative datasets, we choose the C-rate based on the number of nodes in synthetic graph to compare the performance. Take the PPI dataset as the example, C-rate equals 1.00\% means the synthetic graph has 147 nodes compared with original 14755 nodes. For all the synthetic graphs, the nodes number is around 150.

Specifically, we report the performance of various initialization strategies include subgraph sampling( Random, Herding, K-Center) \cite{welling2009herding, farahani2009facility, sener2017active}, and probabilistic label sampling for synthetic $Y'$. After the initialization, the synthetic graph $\mathcal{S}$ will be optimized by gradient matching strategy in GCond.
Then we compare different optimization loss functions, including SoftMarginLoss \cite{cao2019learning} and BCE loss \cite{durand2019learning}, across diverse datasets with pre-choosen condensation rates. Through these experiments, we aim to identify the optimal combination of initialization and loss function that achieves the best trade-off between graph size reduction and model performance in multi-label tasks.


\noindent{\bf Results and Analysis.} 
Finally, by comparing different settings of multi-label scenario on GCond framework, we present the corresponding results of above initialization and optimization with or without structure learning. Specifically,  Table \ref{tab:init} shows the F1-micro scores for four initialization strategies (Random, Herding, K-Center, and Probability) applied to both subgraph and node-only settings. This comparison across datasets allows us to identify which strategies best capture label distributions in multi-label tasks. Then Table \ref{tab:loss1}, \ref{tab:BCEloss} compare performance under SoftMarginLoss and BCELoss optimizations, respectively. Besides, for each initialization method we investigate the performance of synthetic graph with or without structure information. These results help to pinpoint effective combinations of initialization and loss functions.

After analysis of the overall performance with key observations, we report the best setting for multi-label adaptation to better generalization in other methods. By observing the experimental results, we draw three corresponding observation conclusions: 

\noindent{\bf \underline{Observation 1}: K-Center sampling achieves optimal initialization.}

In the GCond framework, K-Center sampling consistently provides the highest F1-micro scores for dense, large-scale datasets such as PPI, Yelp, and OGBN-Proteins in Tables  \ref{tab:init}, \ref{tab:loss1}, \ref{tab:BCEloss}. 

In contrast, simpler strategies like Random and Herding yield better results in datasets with lower density, such as DBLP and PCG. 

Following initialization, we tested SoftMarginLoss and BCELoss for optimization. 
Table \ref{tab:loss1} and Table \ref{tab:BCEloss} are the comparison of different initialization methods optimized by SoftMarginLoss and BCELoss, respectively. Whole dataset indicates the performance of GNN trained by original graph $\mathcal{G}$. (a) and (b) are the comparison of learning condensed graph with or without structure $A'$. We present the combined bar and line chart to show a more intuitive comparison of the differences between various combinations of initialization and optimization methods.
Interestingly, we find the best performance distribution is consistence even using different loss functions. K-Center initialization with or without graph structure optimization gains the best performance using BCELoss. In particular, datasets like PPI and DBLP exhibit notable performance gaps that highlight the advantage of structure learning (per Eq.\ref{eq:structure}) in enhancing multi-label generalization.

\noindent{\bf \underline{Observation 2}: Preserving graph structure could improves performance of condensed large datasets.} 
We examined the effects of learning graph structure versus using nodes alone for label initialization. Results show substantial performance improvements when the graph structure is preserved, especially in large-scale datasets 
While all the experiments are conducted with the comparison of learning synthetic graph with or without structure information, we find out that with the additional $A'$ learning, the condensed graph would perform better in the large-scale datasets.

Before condensation the structure information already shows its effectiveness. For example, in the comparison of initialization methods present in Table \ref{tab:init} PPI with a sampling of the C rate of 1. 00\% and the K-Center, retaining the structure achieves an F1 micro score of 47. 77\%, compared to 36. 60\% with nodes alone. During condensation process, optimization with the structure information learning, same pattern is observed across various methods 
(see Table \ref{tab:loss1}, \ref{tab:BCEloss} for a comparison with or without $A'$ learning process), demonstrating the critical role of structural information.

\noindent{\bf \underline{Observation 3}: BCELoss provides better overall results}
When comparing loss functions, BCELoss generally yields higher F1-micro scores, especially when combined with structure learning. For example, 
In Table \ref{tab:BCEloss}, the DBLP dataset with K-Center sampling and structure learning, BCELoss achieves an F1-micro score of $70.74\%$, substantially outperforming the SoftMarginLoss score of 60.77\%. This trend is consistent across multiple datasets, highlighting BCELoss as the optimal choice for multi-label adaptation in GCond.

Across all initialization settings, the subgraph sampled by the K-Center strategy receives the best performance in the PPI, Yelp and OGBN-Proteins rating by F1-micro score. However, Random and Herding work quite well among the less dense multi-label datasets like DBLP and PCG.
After removing the graph structure, by comparing only the set of nodes of the sampled graphs for label initialization and probability synthetic labels in Table \ref{tab:init}, we find that some of the graph structure could have a factor to impact performance like in PPI with 1.00\% using K-Center, the result of the subgraph could achieve 47.77\% with structure while 36.60\% only with nodes. Similarly, K-Center works best among Yelp, OGBN-Proteins, and PCG with node set. Probability synthetic labels capture less of the distribution of the multi-labels in original dataset and only get comparable performance in PPI. After deploying the initialization methods, we also compare these various settings combined with the loss functions discussed. Following the optimization of the condensation process with or without $A'$, we report the results of GCond with SoftMarginLoss and BCELoss optimization strategies in Table \ref{tab:loss1} and Table \ref{tab:BCEloss}, respectively. 
Finally, we report the the general initialization methods comparison visually and F1-macro score in appendix \ref{appendix}. Interestingly, we find the best performance distribution is consistence even using different loss functions. K-Center initialization with or without graph structure optimization gains the best performance using BCELoss. Noticeably, based on structure learning in Eq.\ref{eq:structure} the performances gap is quite obvious in PPI and DBLP datasets. 

\emph{Consistent with Observations 1 and 2, initializing with K-Center sampling and incorporating structure learning $A'$ could increase the performance of synthetic multi-label graph. As shown in Observation 3, arcoss a variety of datasets, BCELoss performs better than SoftMarginLoss. Particularly when paired with K-Center and structural learning, BCELoss improves the F1-micro score, suggesting it can offer greater label retention and general adaptability. 
Therefore, we adopt the K-Center and BCELoss as the initialization and optimization methods with structure information for multi-label scenario.}

\section{Benchmarking}
\label{sec:2}

\begin{table*}[ht]
\centering
\caption{Graph Condensation Methods Performance Comparison with F1-micro and F1-macro. Performance metrics of the model, with the F1-score represented as a decimal value. "-" means the out-of-memory (OOM) errors.}
\resizebox{\textwidth}{!}{%
\begin{tabular}{cccccccccccccccc}
\hline
\multirow{2}{*}{\bf Dataset} & \multirow{2}{*}{\bf C-rate} & \multicolumn{2}{c}{\bf Coreset (Random)} & \multicolumn{2}{c}{\bf Coreset (Herding)} & \multicolumn{2}{c}{\bf Coreset (K-Center)} & \multicolumn{2}{c}{\bf SGDD} & \multicolumn{2}{c}{\bf GCDM} & \multicolumn{2}{c}{\bf GCond} & \multicolumn{2}{c}{\bf Whole Dataset} \\
    &  & \bf F1-micro & \bf F1-macro & \bf F1-micro & \bf F1-macro & \bf F1-micro & \bf F1-macro & \bf F1-micro & \bf F1-macro & \bf F1-micro & \bf F1-macro & \bf F1-micro & \bf F1-macro & \bf F1-micro & \bf F1-macro \\
\hline
\multirow{3}{*}{\bf PPI }&\bf 0.50\% & 41.92 & 14.69 & 41.16 & 12.76 & 45.65 & 15.54 & 44.99 & 22.58 & 48.28 & 23.37 & \bf 49.63 & \bf 23.16 & \multirow{3}{*}{51.26} & \multirow{3}{*}{30.06} \\
    &\bf 1.00\% & 39.07 & 12.58 & 41.23 & 11.66 & 47.43 & 17.36 & 47.02 & 18.26 & 47.07 & 21.76 & \bf 51.35 &\bf 23.86 & & \\
    &\bf 2.00\% & 41.12 & 14.12 & 39.21 & 10.23 & \bf 52.34 &\bf 26.62 & 46.73 & 21.24 & 47.91 & 21.93 & 51.21 & 24.32 & & \\
\hline
\multirow{3}{*}{\bf PPI-large} &\bf  0.10\% & 44.43 & 14.49 & 43.17 & 13.74 & 41.12 & 18.11 & 49.32 & 18.33 & 47.13 & 37.59 &\bf 51.05 &\bf 27.51 & \multirow{3}{*}{51.61} & \multirow{3}{*}{28.23} \\
    &\bf 0.20\% & 45.15 & 15.80 & 43.39 & 15.91 & 40.88 & 13.00 & 44.82 & 15.78 & 46.26 & 22.24 & \bf 51.99 &\bf 29.95 & & \\
    &\bf 0.40\% & 44.01 & 15.43 & 43.61 & 13.74 & 41.04 & 12.89 & 49.89 & 29.08 & 44.08 & 19.48 & \bf 52.23 &\bf 28.59 & & \\
\hline
\multirow{6}{*}{\bf Yelp} &\bf 0.01\% & 33.73 & 5.55 & 33.63 & 7.07 & \bf 34.20 &\bf 13.30 & - & - & 28.00 & 2.78 & 28.40 & 7.51 & \multirow{6}{*}{37.97} & \multirow{6}{*}{15.18} \\
    &\bf 0.02\% & 33.64 & 5.09 & 32.94 & 5.64 & \bf 35.64 &\bf 12.30 & - & - & 27.84 & 3.76 & 33.94 & 6.22 & & \\
    &\bf 0.04\% & 34.01 & 5.41 & 32.87 & 4.67 &\bf 36.51 &\bf 12.51 & - & - & 26.42 & 2.87 & 32.77 & 5.85 & & \\
    &\bf 0.05\% & 33.74 & 5.86 & 32.83 & 5.17 &\bf 36.63 &\bf 11.30 & - & - & 24.10 & 2.42 & 34.11 & 6.86 & & \\
    &\bf 0.10\% & 33.56 & 5.19 & 32.83 & 5.51 &\bf 36.65 &\bf 9.79 & - & - & 24.82 & 2.95 & 33.70 & 8.17 & & \\
    &\bf 0.20\% & 33.75 & 5.50 & 33.20 & 4.71 &\bf 37.61 &\bf 11.48 & - & - & 28.43 & 3.32 & 32.41 & 12.18 & & \\
\hline
\multirow{5}{*}{\bf DBLP} &\bf 0.20\% & 49.28 & 34.81 & 41.02 & 15.41 & 52.70 & 49.75 & 43.17 & 25.06 & 44.41 & 34.84 &\bf 63.36 &\bf 56.27 & \multirow{5}{*}{87.55} & \multirow{5}{*}{86.39} \\
    &\bf 0.40\% & 52.94 & 44.71 & 42.59 & 31.05 & 55.27 & 50.89 & 45.37 & 43.42 & 45.88 & 26.54 &\bf 68.11 &\bf 57.35 & & \\
    &\bf 0.80\% & 62.08 & 56.00 & 41.11 & 15.63 & 59.77 & 56.90 & 43.48 & 25.42 & 46.66 & 46.20 &\bf 70.74 &\bf 68.44 & & \\
    &\bf 1.60\% & 68.39 & 63.75 & 42.24 & 25.78 & 63.87 & 62.34 & 42.73 & 34.22 & 45.99 & 26.47 &\bf 70.49 &\bf 68.92 & & \\
    &\bf 3.20\% & 71.50 & 68.24 & 42.70 & 22.31 & 67.51 & 64.54 & 43.17 & 25.06 & 47.81 & 28.02 &\bf 70.21 &\bf 67.78 & & \\
\hline
\multirow{3}{*}{\bf OGBN-Proteins}&\bf 0.05\% & 14.40 & 1.70 & 11.29 & 4.25 &\bf 28.74 &\bf 7.83 & - & - & 21.82 & 7.95 & 26.38 & 9.98 & \multirow{3}{*}{10.14} & \multirow{3}{*}{9.16} \\
    &\bf 0.10\% & 15.63 & 2.12 & 15.28 & 5.21 &\bf 31.04 &\bf 10.13 & - & - & 22.18 & 3.07 & 29.40 & 7.47 & & \\
    &\bf 0.20\% & 16.90 & 2.06 & 8.55 & 1.64 & 26.36 & 4.82 & - & - & 21.38 & 8.78 &\bf 29.26 &\bf 7.71 & & \\
\hline
\multirow{3}{*}{\bf PCG} &\bf 2.00\% & 15.72 & 5.57 & 18.98 & 3.03 & 27.44 & 9.09 & 27.93 & 15.32 &\bf 29.48 &\bf 13.22 & 18.84 & 14.21 & \multirow{3}{*}{42.26} & \multirow{3}{*}{31.49} \\
    &\bf 4.00\% & 15.83 & 5.20 & 23.03 & 13.84 & 25.13 & 5.73 & 28.40 & 9.00 & \textbf{32.16} & \textbf{8.31} & 25.58 & 13.32 & & \\
    &\bf 8.00\% & 16.67 & 3.02 & 22.48 & 9.36 & 18.21 & 4.59 &\bf 33.50 &\bf 8.58 & 31.88 & 12.51 & 26.37 & 13.90 & & \\
\hline
\multirow{3}{*}{\bf HumanGo} &\bf 2.00\% & 28.88 & 4.57 & 19.92 & 4.39 & 25.38 & 6.15 & 35.58 & 10.06 &\bf 36.71 &\bf 6.53 & 30.68 & 11.24 & \multirow{3}{*}{51.67} & \multirow{3}{*}{25.57} \\
    &\bf 4.00\% & 27.95 & 7.98 & 31.70 & 6.66 & 31.94 & 7.24 & 35.39 & 9.19 &\bf 35.79 &\bf 7.31 & 34.91 & 10.74 & & \\
    &\bf 8.00\% &\bf 37.57 &\bf 9.58 & 32.44 & 5.50 & 36.50 & 7.67 & 35.25 & 9.74 & 36.54 & 7.35 & 37.18 & 12.05 & & \\
\hline
\multirow{3}{*}{\bf EukaryoteGo} &\bf 1.00\% & 31.45 & 5.57 & 21.60 & 4.24 & 17.79 & 2.66 & 29.40 & 3.25 &\bf 36.93 &\bf 4.09 & 35.24 & 5.57 & \multirow{3}{*}{45.86} & \multirow{3}{*}{12.27} \\
    &\bf 2.00\% & 30.72 & 5.03 & 26.74 & 5.16 & 20.99 & 3.03 & 28.69 & 3.28 &\bf 36.98 &\bf 4.09 & 36.38 & 7.00 & & \\
    &\bf 3.00\% & 30.79 & 4.18 & 25.99 & 3.43 & 22.08 & 4.31 & 36.57 & 4.75 & 36.97 & 4.84 &\bf 38.90 &\bf 6.04 & & \\
\hline
\end{tabular}\label{tab:whole}
}
\end{table*}
In this section, followed the best setting discussed in Section \ref{sec:1}, we adapt all the discussed condensation methods into multi-label scenario based on the former settings. 
The results in Table \ref{tab:whole} illustrate the efficacy of different Coreset selection strategies (Random, Herding, K-Center) and graph condensation methods (SGDD, GCDM, GCond) compared to the full dataset.  The comparison is performed at different condensation rates (C-rate), providing insights into the impact of these methods on classification performance in multi-label tasks.  Specifically, SGDD and GCond both choosing gradient matching for the GNNs trained on original graph $\mathcal{G}$ and synthetic graph $\mathcal{S}$ directly. GCDM uses the embedding matching based on the nodes set belongs to the same class in multi-label.

We have several noteworthy observations concerning the methods, the datasets, the evaluation settings, and the overall results.
We highlight the main observations below. Here, the performance metrics of the model, with the F1-score represented as a decimal value. 

\noindent{\bf \underline{Observation 1}: GCond achieves the best results across datasets.} 
GCond consistently outperforms other methods on most datasets, particularly excelling in large-scale datasets with substantial label complexity. On the PPI dataset, GCond achieves F1-micro and F1-macro scores of 51.35\% and 23.86\%, respectively, at a 1.00\% condensation rate. This robustness extends to PPI-large, where it achieves scores of 52.23\% (F1-micro) and 28.59\% (F1-macro) at a 0.40\% condensation rate.  Similarly, on DBLP, GCond demonstrates its superiority with F1-micro and F1-macro scores of 70.74\% and 68.44\%, respectively, at 0.80\% condensation, showcasing its ability to preserve label correlations and structural integrity during condensation. However, GCond's advantage diminishes in the Yelp dataset at high condensation levels, likely due to the dataset's unique label structure, which poses challenges for methods relying on label coherence preservation.

\noindent{\bf \underline{Observation 2}: K-Center demonstrates stability in extreme condensation setting.}
The K-Center method shows consistent performance, particularly on large-scale and dense-label datasets, highlighting its strength in preserving diversity at low condensation rates. On the Yelp dataset, K-Center outperforms traditional Coreset methods, achieving the highest F1-micro score (37.61\%) at a 0.20\% condensation rate. This suggests that its selection strategy effectively retains essential structural and label information, even under high condensation, making it advantageous for large-scale settings where extreme data reduction risks losing critical information.

\noindent{\bf \underline{Observation 3}: SGDD’s performance is constrained by scalability issues on larger datasets.}
SGDD faces out-of-memory (OOM) errors on large datasets like Yelp and OGBN-Proteins, revealing its scalability limitations. However, on smaller datasets, SGDD performs competitively, benefiting from its structure-broadcasting graphon technique, which efficiently captures structural patterns. For example, on datasets such as PCG, HumanGo, and EukaryoteGo, SGDD achieves strong F1-micro and F1-macro scores, demonstrating its effectiveness on smaller scales. Nevertheless, its applicability to larger datasets requires further optimization.

In conclusion, the efficacy of each condensation method varies depending on the characteristics of the dataset. GCond’s superior performance on complex, high-dimensional datasets such as DBLP and PPI demonstrates its strong alignment with datasets that require maintaining label consistency and intricate relationships. In contrast, simpler methods like Random and Herding exhibit relatively stable performance but are outperformed by GCond and K-Center on datasets where preserving structure and label diversity is critical. These findings underscore the importance of selecting methods based on dataset scale, complexity, and label interaction patterns.

\section{Conclusion}
In this benchmark, we presented a comprehensive framework for multi-label graph condensation, adapting traditional single-label methods to address the complexities inherent in multi-label classification tasks. The results show that the GCond method not only excels in maintaining the structural integrity of the original graph, but also enhances the label correlations, leading to more accurate multi-label predictions.

\bibliography{main}

@article{kipf2016semi,
  title={Semi-supervised classification with graph convolutional networks},
  author={Kipf, Thomas N and Welling, Max},
  journal={arXiv preprint arXiv:1609.02907},
  year={2016}
}

@misc{hamilton2018,
      title={Inductive Representation Learning on Large Graphs}, 
      author={William L. Hamilton and Rex Ying and Jure Leskovec},
      year={2018},
      eprint={1706.02216},
      archivePrefix={arXiv},
      primaryClass={cs.SI},
      url={https://arxiv.org/abs/1706.02216}, 
}

@article{Li2019DeepGCNsCG,
  title={DeepGCNs: Can GCNs Go As Deep As CNNs?},
  author={G. Li and Matthias M{\"u}ller and Ali K. Thabet and Bernard Ghanem},
  journal={2019 IEEE/CVF International Conference on Computer Vision (ICCV)},
  year={2019},
  pages={9266-9275},
  url={https://api.semanticscholar.org/CorpusID:201070021}
}

@article{zhang2021graph,
  title={Graph neural networks and their current applications in bioinformatics},
  author={Zhang, Xiao-Meng and Liang, Li and Liu, Lin and Tang, Ming-Jing},
  journal={Frontiers in genetics},
  volume={12},
  pages={690049},
  year={2021},
  publisher={Frontiers Media SA}
}

@article{Hashemi2024ACS,
  title={A Comprehensive Survey on Graph Reduction: Sparsification, Coarsening, and Condensation},
  author={Mohammad Hashemi and Shengbo Gong and Juntong Ni and Wenqi Fan and B. Aditya Prakash and Wei Jin},
  journal={ArXiv},
  year={2024},
  volume={abs/2402.03358},
  url={https://api.semanticscholar.org/CorpusID:267341687}
}

@article{Loukas2018GraphRB,
  title={Graph reduction by local variation},
  author={Andreas Loukas},
  journal={ArXiv},
  year={2018},
  volume={abs/1808.10650},
  url={https://api.semanticscholar.org/CorpusID:52144738}
}

@article{Jin2022CondensingGV,
  title={Condensing Graphs via One-Step Gradient Matching},
  author={Wei Jin and Xianfeng Tang and Haoming Jiang and Zheng Li and Danqing Zhang and Jiliang Tang and Bin Ying},
  journal={Proceedings of the 28th ACM SIGKDD Conference on Knowledge Discovery and Data Mining},
  year={2022},
  url={https://api.semanticscholar.org/CorpusID:249712265}
}

@misc{GCond,
      title={Graph Condensation for Graph Neural Networks}, 
      author={Wei Jin and Lingxiao Zhao and Shichang Zhang and Yozen Liu and Jiliang Tang and Neil Shah},
      year={2022},
      eprint={2110.07580},
      archivePrefix={arXiv},
      primaryClass={cs.LG},
      url={https://arxiv.org/abs/2110.07580}, 
}

@article{tsoumakas2008multi,
  title={Multi-label classification: An overview},
  author={Tsoumakas, Grigorios and Katakis, Ioannis},
  journal={Data Warehousing and Mining: Concepts, Methodologies, Tools, and Applications},
  pages={64--74},
  year={2008},
  publisher={IGI Global}
}

@misc{GCDM,
      title={Graph Condensation via Receptive Field Distribution Matching}, 
      author={Mengyang Liu and Shanchuan Li and Xinshi Chen and Le Song},
      year={2022},
      eprint={2206.13697},
      archivePrefix={arXiv},
      primaryClass={cs.LG},
      url={https://arxiv.org/abs/2206.13697}, 
}

@misc{SGDD,
      title={Does Graph Distillation See Like Vision Dataset Counterpart?}, 
      author={Beining Yang and Kai Wang and Qingyun Sun and Cheng Ji and Xingcheng Fu and Hao Tang and Yang You and Jianxin Li},
      year={2023},
      eprint={2310.09192},
      archivePrefix={arXiv},
      primaryClass={cs.LG},
      url={https://arxiv.org/abs/2310.09192}, 
}

@inproceedings{chen2019multi,
  title={Multi-label image recognition with graph convolutional networks},
  author={Chen, Zhao-Min and Wei, Xiu-Shen and Wang, Peng and Guo, Yanwen},
  booktitle={Proceedings of the IEEE/CVF conference on computer vision and pattern recognition},
  pages={5177--5186},
  year={2019}
}

@article{shi2019mlne,
  title={MLNE: Multi-label network embedding},
  author={Shi, Min and Tang, Yufei and Zhu, Xingquan},
  journal={IEEE transactions on neural networks and learning systems},
  volume={31},
  number={9},
  pages={3682--3695},
  year={2019},
  publisher={IEEE}
}

@article{shi2020multi,
  title={Multi-label graph convolutional network representation learning},
  author={Shi, Min and Tang, Yufei and Zhu, Xingquan and Liu, Jianxun},
  journal={IEEE Transactions on Big Data},
  volume={8},
  number={5},
  pages={1169--1181},
  year={2020},
  publisher={IEEE}
}

@inproceedings{akujuobi2019collaborative,
  title={Collaborative graph walk for semi-supervised multi-label node classification},
  author={Akujuobi, Uchenna and Yufei, Han and Zhang, Qiannan and Zhang, Xiangliang},
  booktitle={2019 IEEE International Conference on Data Mining (ICDM)},
  pages={1--10},
  year={2019},
  organization={IEEE}
}

@article{zhao2023multi,
  title={Multi-label node classification on graph-structured data},
  author={Zhao, Tianqi and Dong, Ngan Thi and Hanjalic, Alan and Khosla, Megha},
  journal={arXiv preprint arXiv:2304.10398},
  year={2023}
}

@article{xu2024survey,
  title={A survey on graph condensation},
  author={Xu, Hongjia and Zhang, Liangliang and Ma, Yao and Zhou, Sheng and Zheng, Zhuonan and Jiajun, Bu},
  journal={arXiv preprint arXiv:2402.02000},
  year={2024}
}

@article{gao2024graph,
  title={Graph condensation: A survey},
  author={Gao, Xinyi and Yu, Junliang and Jiang, Wei and Chen, Tong and Zhang, Wentao and Yin, Hongzhi},
  journal={arXiv preprint arXiv:2401.11720},
  year={2024}
}

@article{zhang2023survey,
  title={A survey on graph neural network acceleration: Algorithms, systems, and customized hardware},
  author={Zhang, Shichang and Sohrabizadeh, Atefeh and Wan, Cheng and Huang, Zijie and Hu, Ziniu and Wang, Yewen and Cong, Jason and Sun, Yizhou and others},
  journal={arXiv preprint arXiv:2306.14052},
  year={2023}
}

@article{zheng2024structure,
  title={Structure-free graph condensation: From large-scale graphs to condensed graph-free data},
  author={Zheng, Xin and Zhang, Miao and Chen, Chunyang and Nguyen, Quoc Viet Hung and Zhu, Xingquan and Pan, Shirui},
  journal={Advances in Neural Information Processing Systems},
  volume={36},
  year={2024}
}

@inproceedings{liu2023cat,
  title={Cat: Balanced continual graph learning with graph condensation},
  author={Liu, Yilun and Qiu, Ruihong and Huang, Zi},
  booktitle={2023 IEEE International Conference on Data Mining (ICDM)},
  pages={1157--1162},
  year={2023},
  organization={IEEE}
}

@inproceedings{xu2023kernel,
  title={Kernel ridge regression-based graph dataset distillation},
  author={Xu, Zhe and Chen, Yuzhong and Pan, Menghai and Chen, Huiyuan and Das, Mahashweta and Yang, Hao and Tong, Hanghang},
  booktitle={Proceedings of the 29th ACM SIGKDD Conference on Knowledge Discovery and Data Mining},
  pages={2850--2861},
  year={2023}
}

@inproceedings{wang2024fast,
  title={Fast graph condensation with structure-based neural tangent kernel},
  author={Wang, Lin and Fan, Wenqi and Li, Jiatong and Ma, Yao and Li, Qing},
  booktitle={Proceedings of the ACM on Web Conference 2024},
  pages={4439--4448},
  year={2024}
}

@article{zhang2024navigating,
  title={Navigating complexity: Toward lossless graph condensation via expanding window matching},
  author={Zhang, Yuchen and Zhang, Tianle and Wang, Kai and Guo, Ziyao and Liang, Yuxuan and Bresson, Xavier and Jin, Wei and You, Yang},
  journal={arXiv preprint arXiv:2402.05011},
  year={2024}
}

@article{wang2018dataset,
  title={Dataset distillation},
  author={Wang, Tongzhou and Zhu, Jun-Yan and Torralba, Antonio and Efros, Alexei A},
  journal={arXiv preprint arXiv:1811.10959},
  year={2018}
}

@article{gao2019graphon,
  title={Graphon control of large-scale networks of linear systems},
  author={Gao, Shuang and Caines, Peter E},
  journal={IEEE Transactions on Automatic Control},
  volume={65},
  number={10},
  pages={4090--4105},
  year={2019},
  publisher={IEEE}
}

@article{ruiz2020graphon,
  title={Graphon neural networks and the transferability of graph neural networks},
  author={Ruiz, Luana and Chamon, Luiz and Ribeiro, Alejandro},
  journal={Advances in Neural Information Processing Systems},
  volume={33},
  pages={1702--1712},
  year={2020}
}

@inproceedings{xia2023implicit,
  title={Implicit graphon neural representation},
  author={Xia, Xinyue and Mishne, Gal and Wang, Yusu},
  booktitle={International Conference on Artificial Intelligence and Statistics},
  pages={10619--10634},
  year={2023},
  organization={PMLR}
}

@inproceedings{pfeiffer2014attributed,
  title={Attributed graph models: Modeling network structure with correlated attributes},
  author={Pfeiffer III, Joseph J and Moreno, Sebastian and La Fond, Timothy and Neville, Jennifer and Gallagher, Brian},
  booktitle={Proceedings of the 23rd international conference on World wide web},
  pages={831--842},
  year={2014}
}

@article{shalizi2011homophily,
  title={Homophily and contagion are generically confounded in observational social network studies},
  author={Shalizi, Cosma Rohilla and Thomas, Andrew C},
  journal={Sociological methods \& research},
  volume={40},
  number={2},
  pages={211--239},
  year={2011},
  publisher={Sage Publications Sage CA: Los Angeles, CA}
}

@article{gutman2006laplacian,
  title={Laplacian energy of a graph},
  author={Gutman, Ivan and Zhou, Bo},
  journal={Linear Algebra and its applications},
  volume={414},
  number={1},
  pages={29--37},
  year={2006},
  publisher={Elsevier}
}

@article{das2016distribution,
  title={Distribution of Laplacian eigenvalues of graphs},
  author={Das, Kinkar Ch and Mojallal, Seyed Ahmad and Trevisan, Vilmar},
  journal={Linear Algebra and its Applications},
  volume={508},
  pages={48--61},
  year={2016},
  publisher={Elsevier}
}

@inproceedings{tang2022rethinking,
  title={Rethinking graph neural networks for anomaly detection},
  author={Tang, Jianheng and Li, Jiajin and Gao, Ziqi and Li, Jia},
  booktitle={International Conference on Machine Learning},
  pages={21076--21089},
  year={2022},
  organization={PMLR}
}

@inproceedings{zhao2023dataset,
  title={Dataset condensation with distribution matching},
  author={Zhao, Bo and Bilen, Hakan},
  booktitle={Proceedings of the IEEE/CVF Winter Conference on Applications of Computer Vision},
  pages={6514--6523},
  year={2023}
}

@inproceedings{durand2019learning,
  title={Learning a deep convnet for multi-label classification with partial labels},
  author={Durand, Thibaut and Mehrasa, Nazanin and Mori, Greg},
  booktitle={Proceedings of the IEEE/CVF conference on computer vision and pattern recognition},
  pages={647--657},
  year={2019}
}

@inproceedings{welling2009herding,
  title={Herding dynamical weights to learn},
  author={Welling, Max},
  booktitle={Proceedings of the 26th annual international conference on machine learning},
  pages={1121--1128},
  year={2009}
}

@book{farahani2009facility,
  title={Facility location: concepts, models, algorithms and case studies},
  author={Farahani, Reza Zanjirani and Hekmatfar, Masoud},
  year={2009},
  publisher={Springer Science \& Business Media}
}

@article{sener2017active,
  title={Active learning for convolutional neural networks: A core-set approach},
  author={Sener, Ozan and Savarese, Silvio},
  journal={arXiv preprint arXiv:1708.00489},
  year={2017}
}

@article{battaglia2018relational,
  title={Relational inductive biases, deep learning, and graph networks},
  author={Battaglia, Peter W and Hamrick, Jessica B and Bapst, Victor and Sanchez-Gonzalez, Alvaro and Zambaldi, Vinicius and Malinowski, Mateusz and Tacchetti, Andrea and Raposo, David and Santoro, Adam and Faulkner, Ryan and others},
  journal={arXiv preprint arXiv:1806.01261},
  year={2018}
}

@article{wu2020comprehensive,
  title={A comprehensive survey on graph neural networks},
  author={Wu, Zonghan and Pan, Shirui and Chen, Fengwen and Long, Guodong and Zhang, Chengqi and Philip, S Yu},
  journal={IEEE transactions on neural networks and learning systems},
  volume={32},
  number={1},
  pages={4--24},
  year={2020},
  publisher={IEEE}
}

@article{zhou2020graph,
  title={Graph neural networks: A review of methods and applications},
  author={Zhou, Jie and Cui, Ganqu and Hu, Shengding and Zhang, Zhengyan and Yang, Cheng and Liu, Zhiyuan and Wang, Lifeng and Li, Changcheng and Sun, Maosong},
  journal={AI open},
  volume={1},
  pages={57--81},
  year={2020},
  publisher={Elsevier}
}

@article{wu2022graph,
  title={Graph neural networks in recommender systems: a survey},
  author={Wu, Shiwen and Sun, Fei and Zhang, Wentao and Xie, Xu and Cui, Bin},
  journal={ACM Computing Surveys},
  volume={55},
  number={5},
  pages={1--37},
  year={2022},
  publisher={ACM New York, NY}
}

@article{hamilton2017inductive,
  title={Inductive representation learning on large graphs},
  author={Hamilton, Will and Ying, Zhitao and Leskovec, Jure},
  journal={Advances in neural information processing systems},
  volume={30},
  year={2017}
}

@article{velickovic2017graph,
  title={Graph attention networks},
  author={Velickovic, Petar and Cucurull, Guillem and Casanova, Arantxa and Romero, Adriana and Lio, Pietro and Bengio, Yoshua and others},
  journal={stat},
  volume={1050},
  number={20},
  pages={10--48550},
  year={2017}
}

@article{zhang2019graph,
  title={Graph convolutional networks: a comprehensive review},
  author={Zhang, Si and Tong, Hanghang and Xu, Jiejun and Maciejewski, Ross},
  journal={Computational Social Networks},
  volume={6},
  number={1},
  pages={1--23},
  year={2019},
  publisher={Springer}
}

@inproceedings{gao2018large,
  title={Large-scale learnable graph convolutional networks},
  author={Gao, Hongyang and Wang, Zhengyang and Ji, Shuiwang},
  booktitle={Proceedings of the 24th ACM SIGKDD international conference on knowledge discovery \& data mining},
  pages={1416--1424},
  year={2018}
}

@inproceedings{bojchevski2020scaling,
  title={Scaling graph neural networks with approximate pagerank},
  author={Bojchevski, Aleksandar and Gasteiger, Johannes and Perozzi, Bryan and Kapoor, Amol and Blais, Martin and R{\'o}zemberczki, Benedek and Lukasik, Michal and G{\"u}nnemann, Stephan},
  booktitle={Proceedings of the 26th ACM SIGKDD International Conference on Knowledge Discovery \& Data Mining},
  pages={2464--2473},
  year={2020}
}

@article{read2011classifier,
  title={Classifier chains for multi-label classification},
  author={Read, Jesse and Pfahringer, Bernhard and Holmes, Geoff and Frank, Eibe},
  journal={Machine learning},
  volume={85},
  pages={333--359},
  year={2011},
  publisher={Springer}
}

@inproceedings{huang2012multi,
  title={Multi-label learning by exploiting label correlations locally},
  author={Huang, Sheng-Jun and Zhou, Zhi-Hua},
  booktitle={Proceedings of the AAAI Conference on Artificial Intelligence},
  volume={26},
  number={1},
  pages={949--955},
  year={2012}
}

@inproceedings{mao2021ultragcn,
  title={UltraGCN: ultra simplification of graph convolutional networks for recommendation},
  author={Mao, Kelong and Zhu, Jieming and Xiao, Xi and Lu, Biao and Wang, Zhaowei and He, Xiuqiang},
  booktitle={Proceedings of the 30th ACM international conference on information \& knowledge management},
  pages={1253--1262},
  year={2021}
}

@article{zhang2019star,
  title={Star-gcn: Stacked and reconstructed graph convolutional networks for recommender systems},
  author={Zhang, Jiani and Shi, Xingjian and Zhao, Shenglin and King, Irwin},
  journal={arXiv preprint arXiv:1905.13129},
  year={2019}
}

@article{jiang2021could,
  title={Could graph neural networks learn better molecular representation for drug discovery? A comparison study of descriptor-based and graph-based models},
  author={Jiang, Dejun and Wu, Zhenxing and Hsieh, Chang-Yu and Chen, Guangyong and Liao, Ben and Wang, Zhe and Shen, Chao and Cao, Dongsheng and Wu, Jian and Hou, Tingjun},
  journal={Journal of cheminformatics},
  volume={13},
  pages={1--23},
  year={2021},
  publisher={Springer}
}

@article{duvenaud2015convolutional,
  title={Convolutional networks on graphs for learning molecular fingerprints},
  author={Duvenaud, David K and Maclaurin, Dougal and Iparraguirre, Jorge and Bombarell, Rafael and Hirzel, Timothy and Aspuru-Guzik, Al{\'a}n and Adams, Ryan P},
  journal={Advances in neural information processing systems},
  volume={28},
  year={2015}
}

@inproceedings{dou2020enhancing,
  title={Enhancing graph neural network-based fraud detectors against camouflaged fraudsters},
  author={Dou, Yingtong and Liu, Zhiwei and Sun, Li and Deng, Yutong and Peng, Hao and Yu, Philip S},
  booktitle={Proceedings of the 29th ACM international conference on information \& knowledge management},
  pages={315--324},
  year={2020}
}

@article{wang2021review,
  title={A review on graph neural network methods in financial applications},
  author={Wang, Jianian and Zhang, Sheng and Xiao, Yanghua and Song, Rui},
  journal={arXiv preprint arXiv:2111.15367},
  year={2021}
}

@article{jiang2022graph,
  title={Graph neural network for traffic forecasting: A survey},
  author={Jiang, Weiwei and Luo, Jiayun},
  journal={Expert systems with applications},
  volume={207},
  pages={117921},
  year={2022},
  publisher={Elsevier}
}

@article{xu2018powerful,
  title={How powerful are graph neural networks?},
  author={Xu, Keyulu and Hu, Weihua and Leskovec, Jure and Jegelka, Stefanie},
  journal={arXiv preprint arXiv:1810.00826},
  year={2018}
}

@article{ching2015one,
  title={One trillion edges: Graph processing at facebook-scale},
  author={Ching, Avery and Edunov, Sergey and Kabiljo, Maja and Logothetis, Dionysios and Muthukrishnan, Sambavi},
  journal={Proceedings of the VLDB Endowment},
  volume={8},
  number={12},
  pages={1804--1815},
  year={2015},
  publisher={VLDB Endowment}
}

@article{hasan2011survey,
  title={A survey of link prediction in social networks},
  author={Hasan, Mohammad Al and Zaki, Mohammed J},
  journal={Social network data analytics},
  pages={243--275},
  year={2011},
  publisher={Springer}
}

@article{xiao2022graph,
  title={Graph neural networks in node classification: survey and evaluation},
  author={Xiao, Shunxin and Wang, Shiping and Dai, Yuanfei and Guo, Wenzhong},
  journal={Machine Vision and Applications},
  volume={33},
  number={1},
  pages={4},
  year={2022},
  publisher={Springer}
}

@inproceedings{zhang2018end,
  title={An end-to-end deep learning architecture for graph classification},
  author={Zhang, Muhan and Cui, Zhicheng and Neumann, Marion and Chen, Yixin},
  booktitle={Proceedings of the AAAI conference on artificial intelligence},
  volume={32},
  number={1},
  year={2018}
}

@article{chou2007euk,
  title={Euk-mPLoc: a fusion classifier for large-scale eukaryotic protein subcellular location prediction by incorporating multiple sites},
  author={Chou, Kuo-Chen and Shen, Hong-Bin},
  journal={Journal of Proteome Research},
  volume={6},
  number={5},
  pages={1728--1734},
  year={2007},
  publisher={ACS Publications}
}

@article{hu2020ogb,
  title={Open Graph Benchmark: Datasets for Machine Learning on Graphs},
  author={Hu, Weihua and Fey, Matthias and Zitnik, Marinka and Dong, Yuxiao and Ren, Hongyu and Liu, Bowen and Catasta, Michele and Leskovec, Jure},
  journal={arXiv preprint arXiv:2005.00687},
  year={2020}
}

@article{zeng2019graphsaint,
  title={Graphsaint: Graph sampling based inductive learning method},
  author={Zeng, Hanqing and Zhou, Hongkuan and Srivastava, Ajitesh and Kannan, Rajgopal and Prasanna, Viktor},
  journal={arXiv preprint arXiv:1907.04931},
  year={2019}
}

@article{church2017word2vec,
  title={Word2Vec},
  author={Church, Kenneth Ward},
  journal={Natural Language Engineering},
  volume={23},
  number={1},
  pages={155--162},
  year={2017},
  publisher={Cambridge University Press}
}

@article{liberzon2015molecular,
  title={The molecular signatures database hallmark gene set collection},
  author={Liberzon, Arthur and Birger, Chet and Thorvaldsd{\'o}ttir, Helga and Ghandi, Mahmoud and Mesirov, Jill P and Tamayo, Pablo},
  journal={Cell systems},
  volume={1},
  number={6},
  pages={417--425},
  year={2015},
  publisher={Elsevier}
}

@article{huntley2015goa,
  title={The GOA database: gene ontology annotation updates for 2015},
  author={Huntley, Rachael P and Sawford, Tony and Mutowo-Meullenet, Prudence and Shypitsyna, Aleksandra and Bonilla, Carlos and Martin, Maria J and O'Donovan, Claire},
  journal={Nucleic acids research},
  volume={43},
  number={D1},
  pages={D1057--D1063},
  year={2015},
  publisher={Oxford University Press}
}

@article{yang2020prediction,
  title={Prediction of human-virus protein-protein interactions through a sequence embedding-based machine learning method},
  author={Yang, Xiaodi and Yang, Shiping and Li, Qinmengge and Wuchty, Stefan and Zhang, Ziding},
  journal={Computational and structural biotechnology journal},
  volume={18},
  pages={153--161},
  year={2020},
  publisher={Elsevier}
}

@article{pinero2020disgenet,
  title={The DisGeNET knowledge platform for disease genomics: 2019 update},
  author={Pi{\~n}ero, Janet and Ram{\'\i}rez-Anguita, Juan Manuel and Sa{\"u}ch-Pitarch, Josep and Ronzano, Francesco and Centeno, Emilio and Sanz, Ferran and Furlong, Laura I},
  journal={Nucleic acids research},
  volume={48},
  number={D1},
  pages={D845--D855},
  year={2020},
  publisher={Oxford University Press}
}

@article{bhattacharya2011mesh,
  title={MeSH: a window into full text for document summarization},
  author={Bhattacharya, Sanmitra and Ha- Thuc, Viet and Srinivasan, Padmini},
  journal={Bioinformatics},
  volume={27},
  number={13},
  pages={i120--i128},
  year={2011},
  publisher={Oxford University Press}
}

@inproceedings{gao2021graph,
  title={Graph neural architecture search},
  author={Gao, Yang and Yang, Hong and Zhang, Peng and Zhou, Chuan and Hu, Yue},
  booktitle={International joint conference on artificial intelligence},
  year={2021},
  organization={International Joint Conference on Artificial Intelligence}
}

@article{cao2019learning,
  title={Learning imbalanced datasets with label-distribution-aware margin loss},
  author={Cao, Kaidi and Wei, Colin and Gaidon, Adrien and Arechiga, Nikos and Ma, Tengyu},
  journal={Advances in neural information processing systems},
  volume={32},
  year={2019}
}

@article{ding2022data,
  title={Data augmentation for deep graph learning: A survey},
  author={Ding, Kaize and Xu, Zhe and Tong, Hanghang and Liu, Huan},
  journal={ACM SIGKDD Explorations Newsletter},
  volume={24},
  number={2},
  pages={61--77},
  year={2022},
  publisher={ACM New York, NY, USA}
}

@article{zhu2021survey,
  title={A survey on graph structure learning: Progress and opportunities},
  author={Zhu, Yanqiao and Xu, Weizhi and Zhang, Jinghao and Du, Yuanqi and Zhang, Jieyu and Liu, Qiang and Yang, Carl and Wu, Shu},
  journal={arXiv preprint arXiv:2103.03036},
  year={2021}
}
\bibliographystyle{tmlr}

\appendix
\section{Appendix}
\label{appendix}

\subsection{Adapted GCond Algorithm}
\begin{algorithm}[ht]
    \caption{Multi-Label GCond Adaptation}
    \label{alg:multi_label_gcond}
    \DontPrintSemicolon
    \KwIn{Training data $ \mathcal{G} = (A, X, Y) $}
    Initialize multi-label synthetic graph $\mathcal{S}$ with ${X', Y'}$ \;
    \For{ $ k = 0 $ \KwTo $ K - 1 $} {
        Initialize $ \theta_0 \sim P_{\theta_0} $ \;
        \For{ $ t = 0 $ \KwTo $ T - 1 $} {
            $ D' \leftarrow 0 $ \;
            Compute $ A' = g_{\phi}(X') $, then $ \mathcal{S} = (A', X', Y') $ \;
            Compute $ \mathcal{L}^\mathcal{G} = \mathcal{L}(GNN_{\theta_t^{\mathcal{G}}}(A, X), Y) $ \;
            Compute $ \mathcal{L}^\mathcal{S} = \mathcal{L}(GNN_{\theta_t^{\mathcal{S}}}(A', X', Y')) $ \;
            $ D' \leftarrow D' + D(\nabla_{\theta_t} \mathcal{L}^\mathcal{G}, \nabla_{\theta_t} \mathcal{L}^\mathcal{S}) $ \;

            \If{ $ t \% (\tau_1 + \tau_2) < \tau_1 $} {
                Update $ X' \leftarrow X' - \eta_1 \nabla_{X'} D' $ \;
            }
            \Else {
                Update $ \phi \leftarrow \phi - \eta_2 \nabla_{\phi} D' $ \;
            }
            Update $ \theta_{t+1} \leftarrow \text{opt}_{\theta} (\theta_t, \mathcal{S}, \tau_{\theta}) $ \;
        }
        Compute new adjacency matrix: $ A' = g_{\phi}(X') $ \;
        \For{each edge $ (i,j) $ in $ A' $} {
            $ A'_{i,j} \leftarrow A'_{i,j} $ if $ A'_{i,j} > \delta $, otherwise $ A'_{i,j} \leftarrow 0 $ \;
        }
    }
    \KwRet{ $ (A', X', Y') $ }
\end{algorithm}

\subsection{Visualization of Multi-Label Datasets}

In this section, we delve deeper into the characteristics of the condensed graphs by analyzing class distribution and label correlation within the multi-label datasets. This detailed examination is crucial for understanding how effectively the condensation methods preserve the multi-label relationships present in the original datasets.

After obtaining the full benchmark results, we provide a detailed analysis of the condensed graphs by examining the class distribution and label correlation within the synthetic graphs. These statistical properties are crucial for understanding how well the condensed graphs preserve the multi-label relationships present in the original datasets.
\begin{itemize}
    \item Label Correlation: 
    To analyze the relationships between labels in the dataset, we define the correlation matrix:
    Let \( G_{\text{labels}} \) be the set of labels from the original dataset \( \mathcal{G} \), and \( Y \) be the corresponding original label matrix. We denote \( M(i, j) \) as the occurrence times of labels \( L_i \) and \( L_j \), and \( N_i \) as the total number of times label \( L_i \) appears.
    
    The conditional probability \( P(i, j) \) is defined as:
    \begin{equation}
    P(i, j) = \frac{M(i, j)}{N_i}
    \end{equation}
    
    This conditional probability matrix \( P \) models the edges as a co-occurrence matrix. 
    
    Next, we define the diagonal matrix \( D \) as follows:
    \begin{equation}
    D = \text{Diag}(P)
    \end{equation}
    
    Finally, the Laplacian matrix \( L_o \) is computed as:
    \begin{equation}
    L_o = D - P
    \end{equation}
    By computing the label correlation matrix, we investigate whether the co-occurrence patterns between labels in the original graph are retained in the condensed graph.
    \item Class Distribution: We analyze how evenly or unevenly the labels are distributed across the classes in different datasets by index, which helps assess the quality of the condensation process.
\end{itemize}

\begin{figure*}[ht]
    \centering
    \begin{subfigure}[b]{0.24\textwidth}
        \centering
        \includegraphics[width=\textwidth]{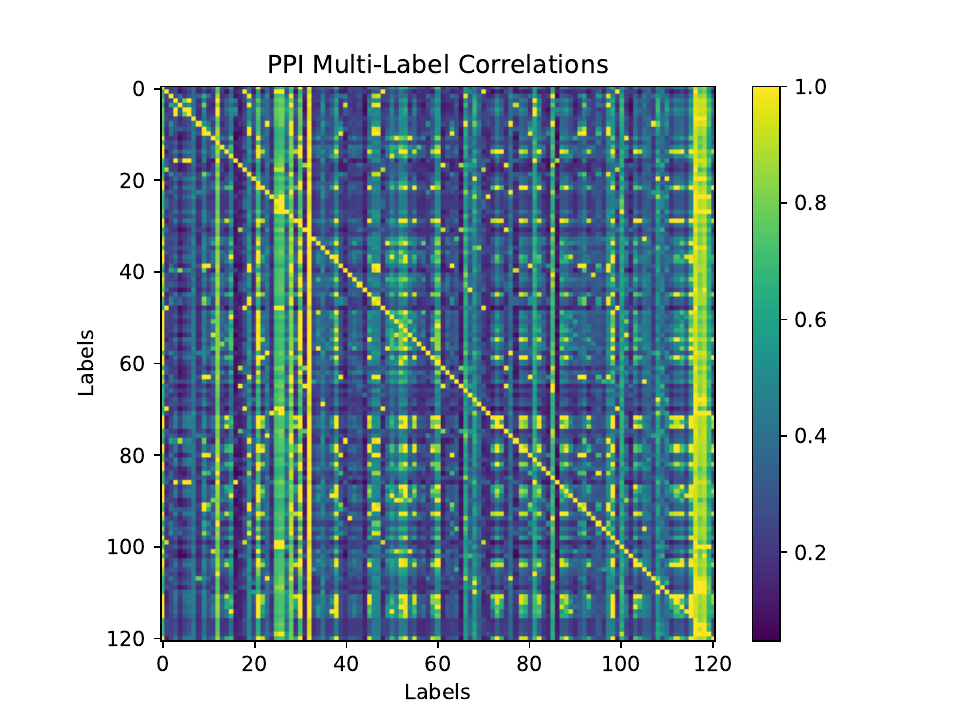}
        \caption{PPI}
    \end{subfigure}
    \begin{subfigure}[b]{0.24\textwidth}
        \centering
        \includegraphics[width=\textwidth]{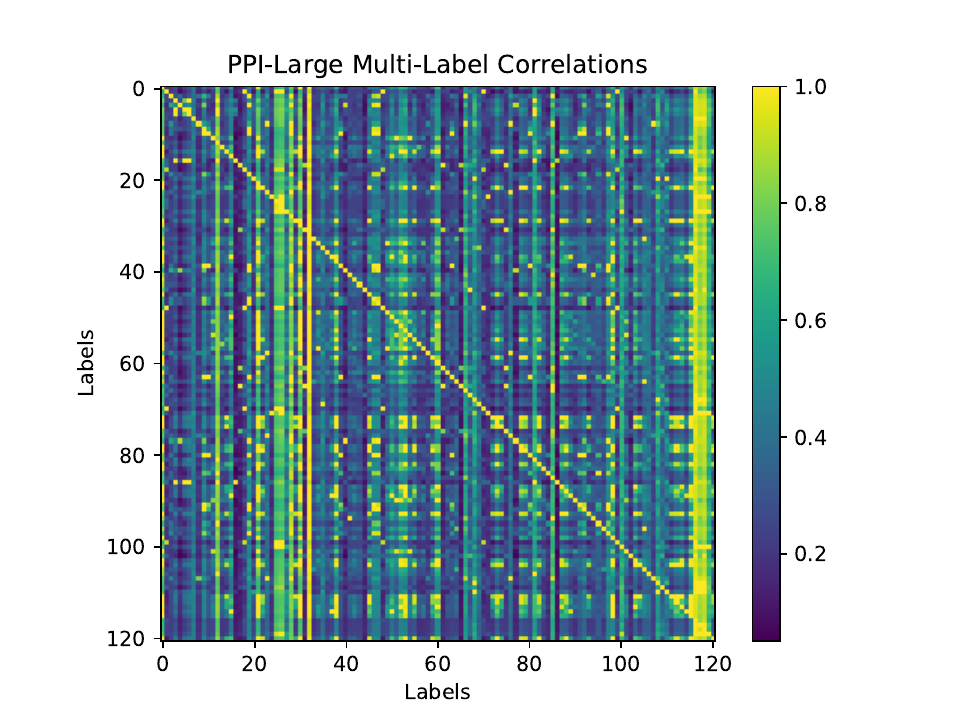}
        \caption{PPI-Large}
    \end{subfigure}
    \begin{subfigure}[b]{0.24\textwidth}
        \centering
        \includegraphics[width=\textwidth]{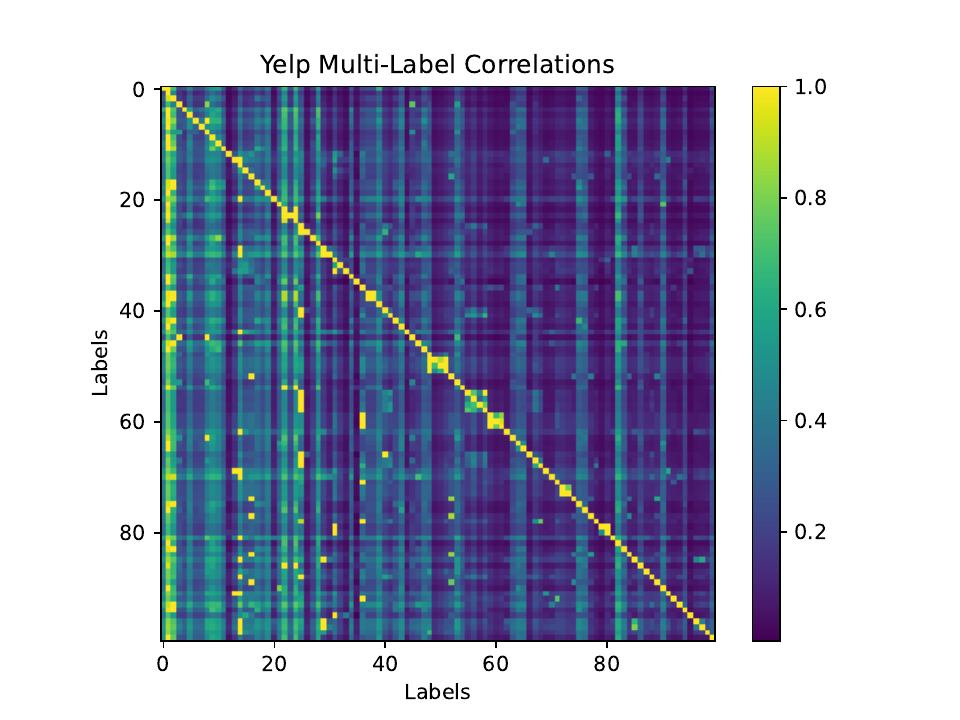}
        \caption{Yelp}
    \end{subfigure}
    \begin{subfigure}[b]{0.24\textwidth}
        \centering
        \includegraphics[width=\textwidth]{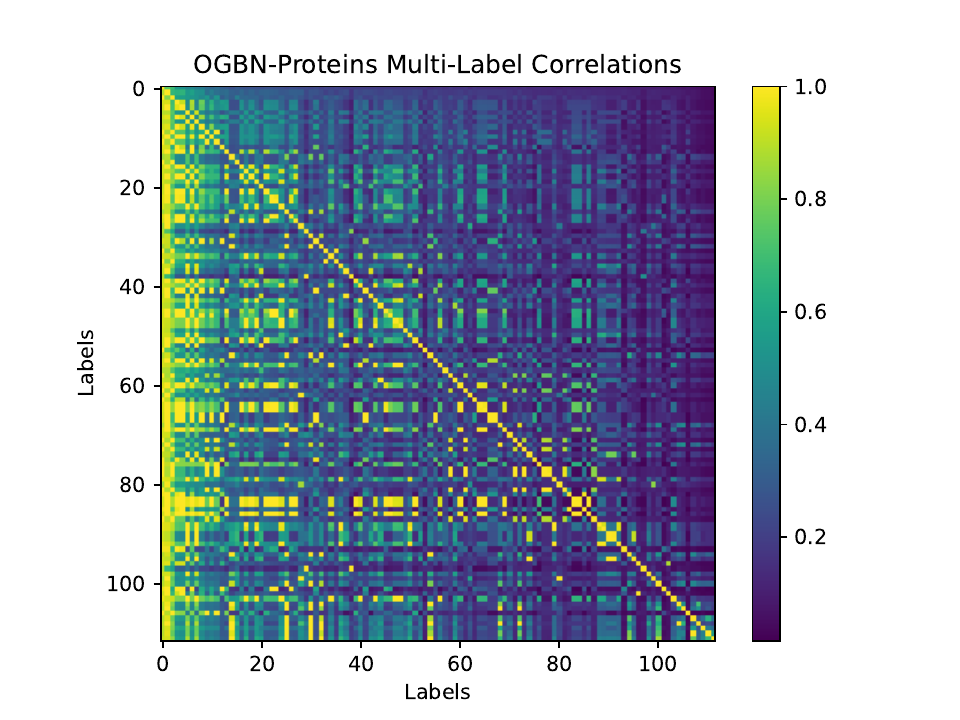}
        \caption{OGNB-Proteins}
    \end{subfigure}

    \vskip\baselineskip 

    \begin{subfigure}[b]{0.24\textwidth}
        \centering
        \includegraphics[width=\textwidth]{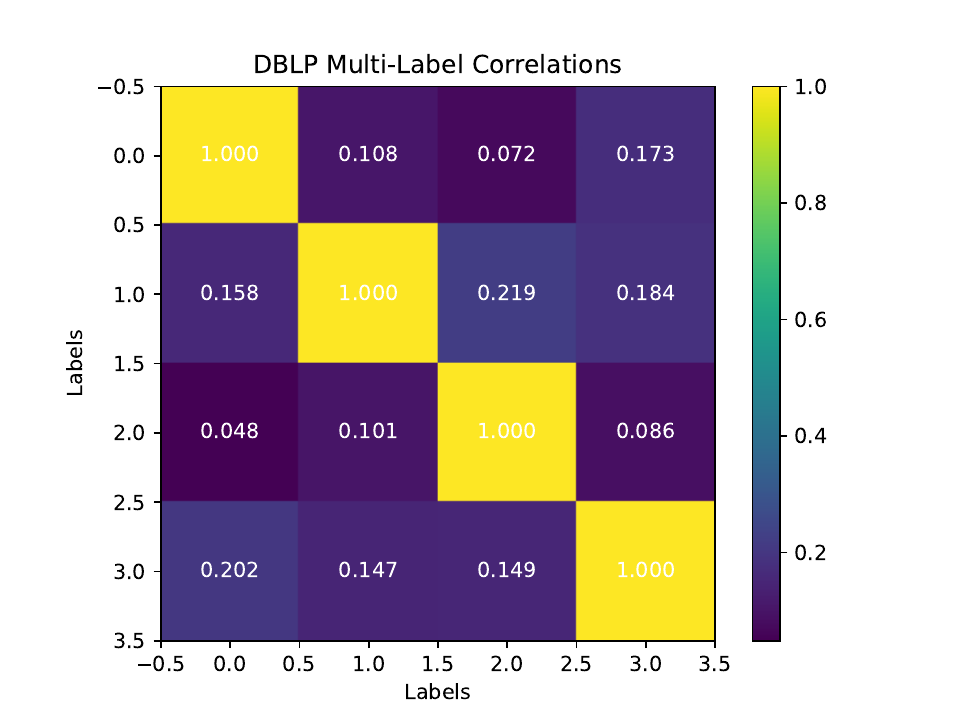}
        \caption{DBLP}
    \end{subfigure}
    \begin{subfigure}[b]{0.24\textwidth}
        \centering
        \includegraphics[width=\textwidth]{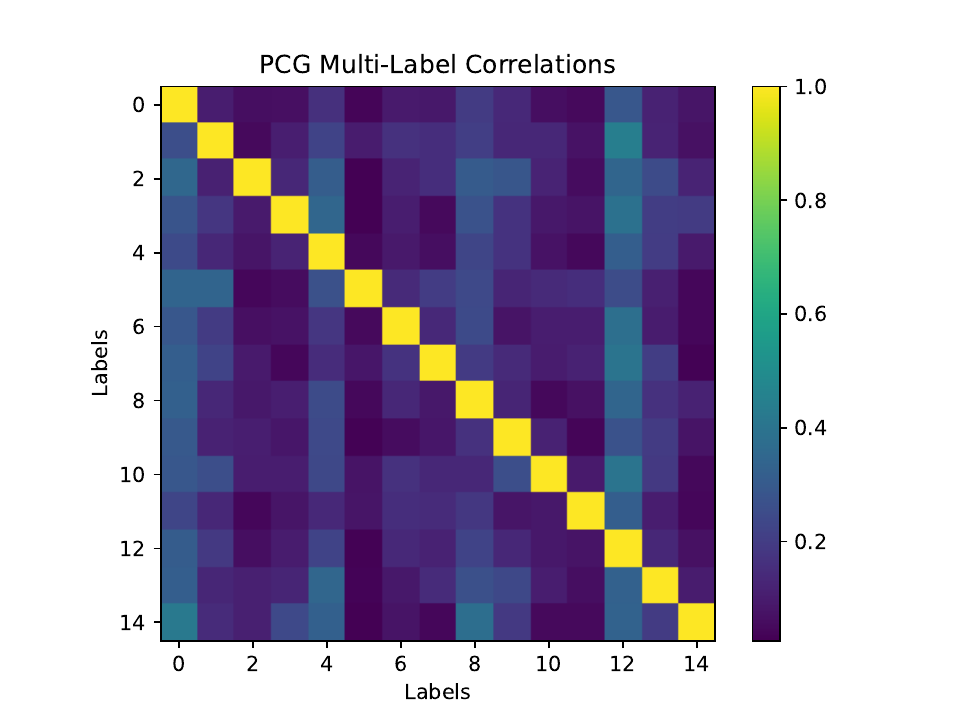}
        \caption{PCG}
    \end{subfigure}
    \begin{subfigure}[b]{0.24\textwidth}
        \centering
        \includegraphics[width=\textwidth]{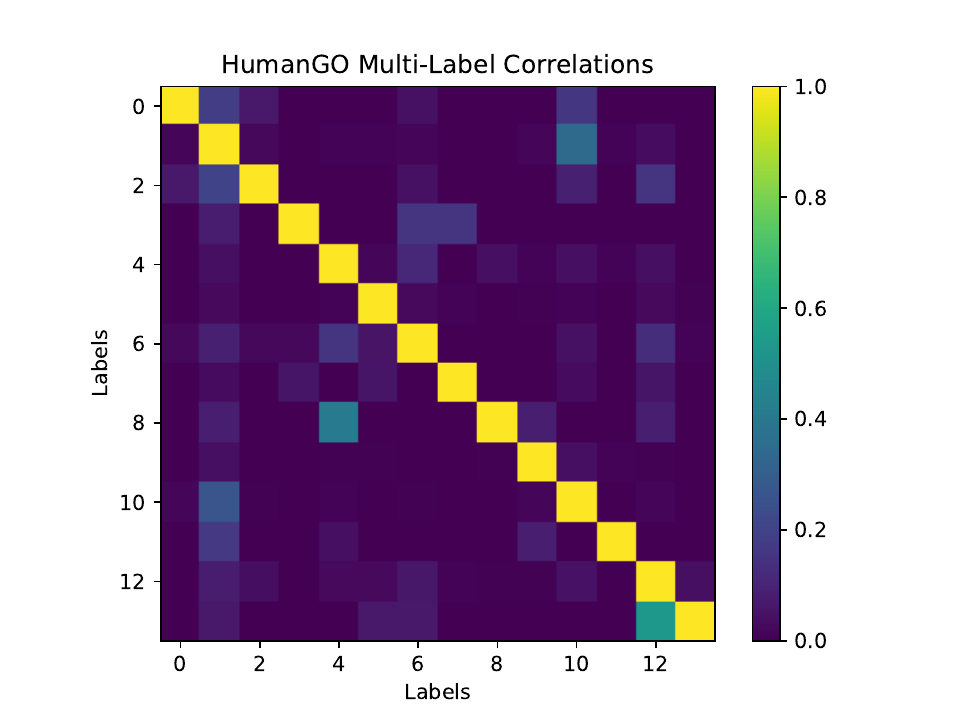}
        \caption{HumanGO}
    \end{subfigure}
    \begin{subfigure}[b]{0.24\textwidth}
        \centering
        \includegraphics[width=\textwidth]{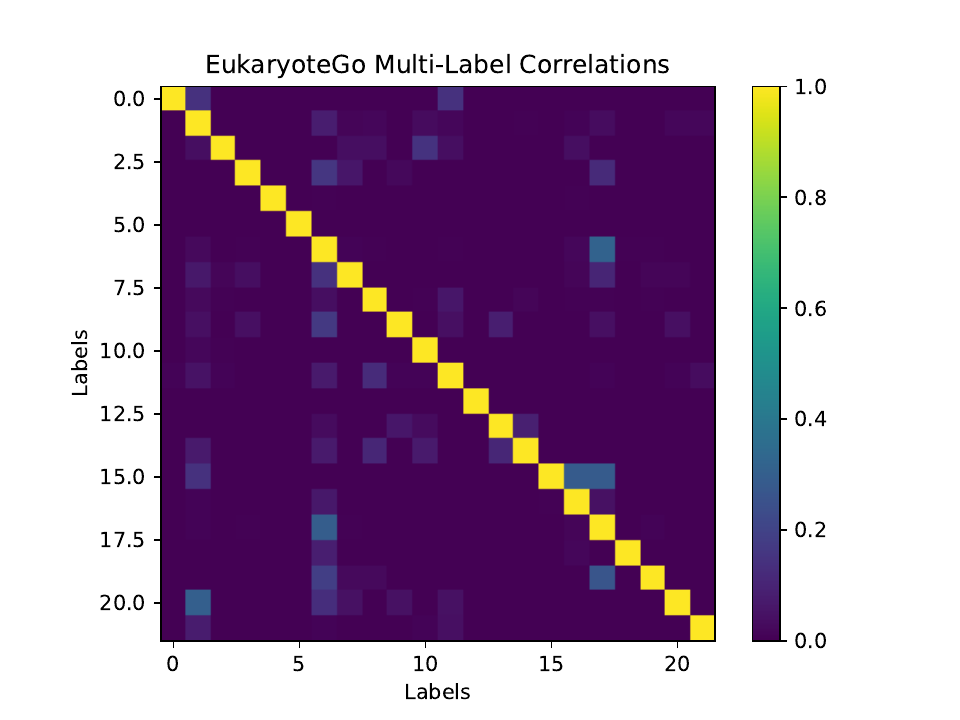}
        \caption{EukaryoteGo}
    \end{subfigure}
    
    \caption{Multi-label Correlation Visualization}
    \label{ap:label}
\end{figure*}

\begin{figure*}[ht]
    \centering
    \begin{subfigure}[b]{0.24\textwidth}
        \centering
        \includegraphics[width=\textwidth]{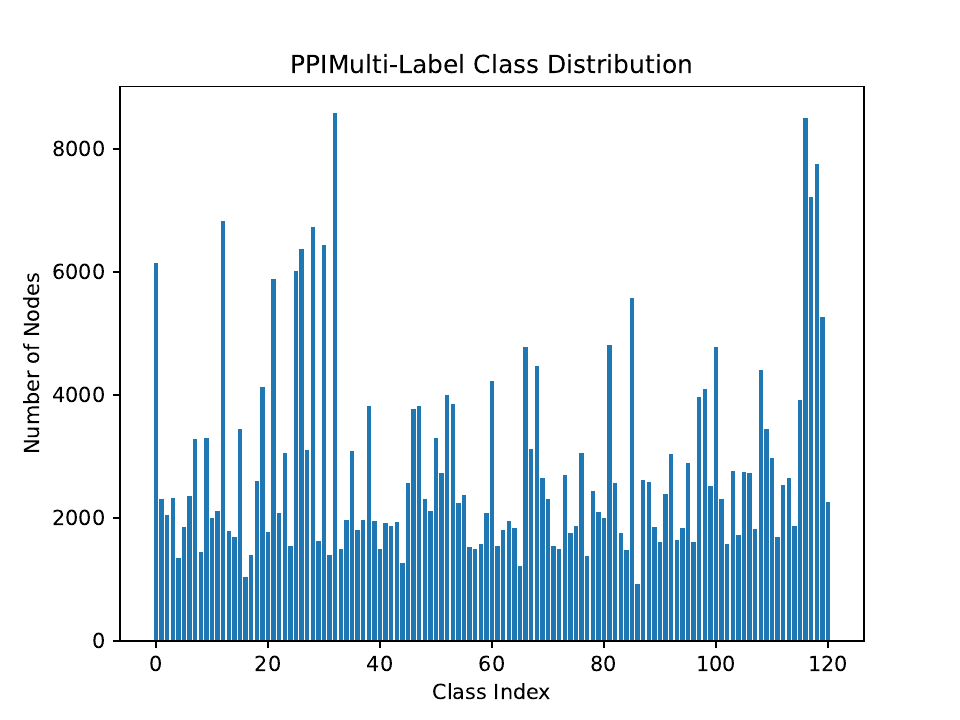}
        \caption{PPI }
    \end{subfigure}
    \begin{subfigure}[b]{0.24\textwidth}
        \centering
        \includegraphics[width=\textwidth]{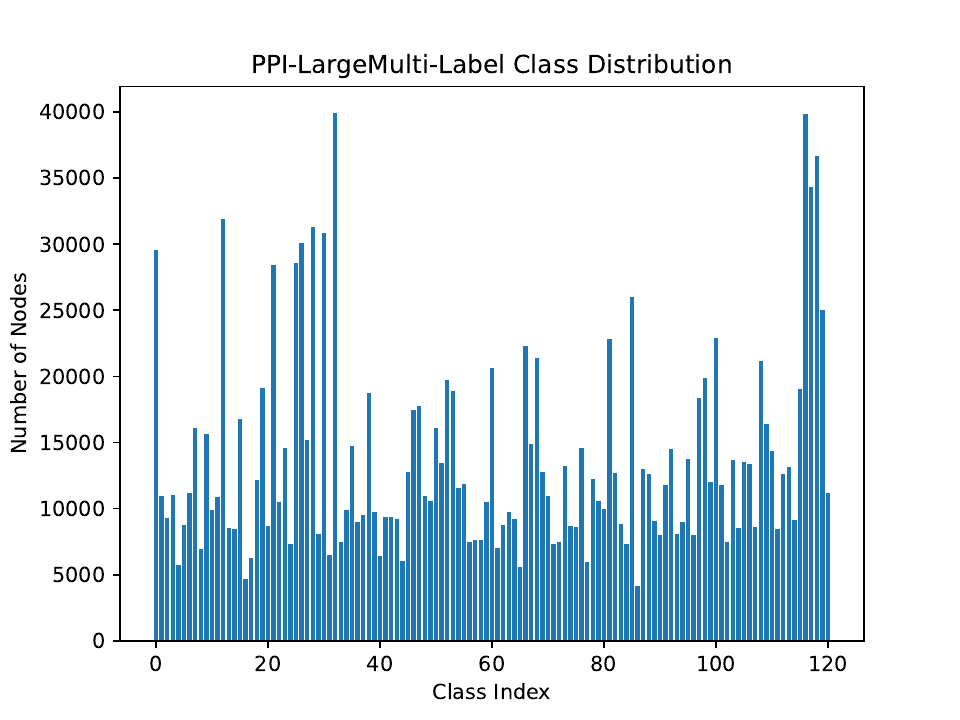}
        \caption{PPI-Large }
    \end{subfigure}
    \begin{subfigure}[b]{0.24\textwidth}
        \centering
        \includegraphics[width=\textwidth]{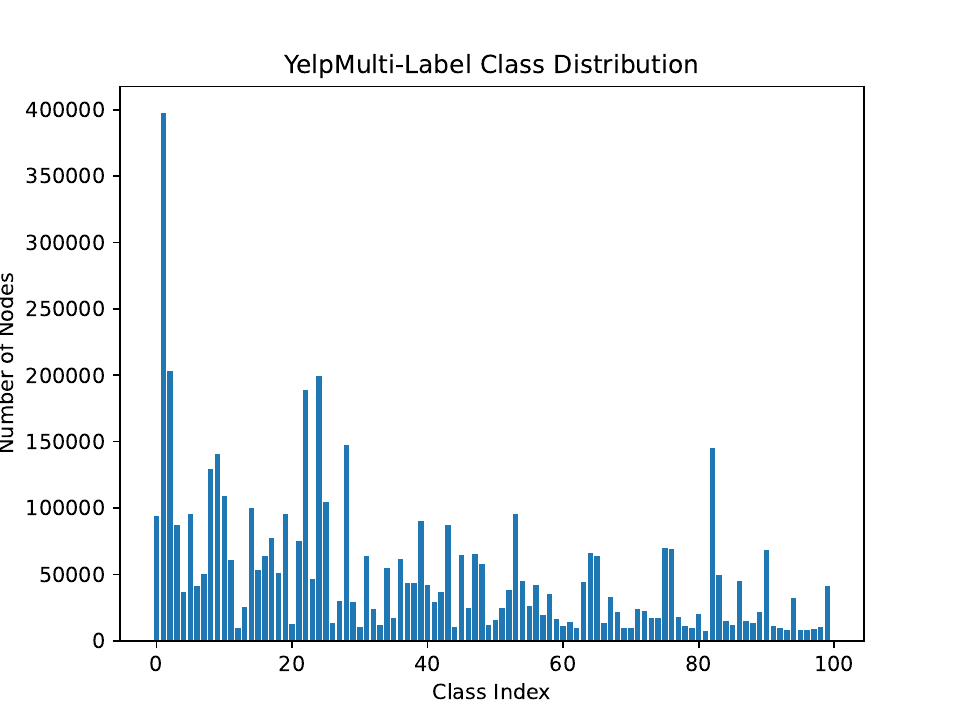}
        \caption{Yelp }
    \end{subfigure}
    \begin{subfigure}[b]{0.24\textwidth}
        \centering
        \includegraphics[width=\textwidth]{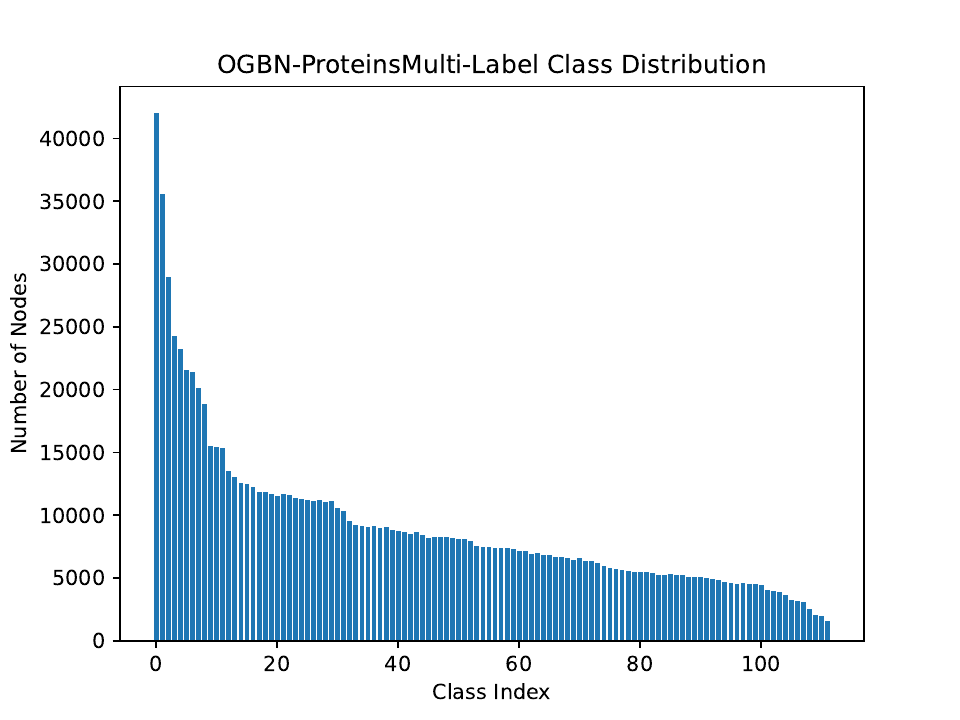}
        \caption{OGBN-Proteins }
    \end{subfigure}

    \vskip\baselineskip 

    \begin{subfigure}[b]{0.24\textwidth}
        \centering
        \includegraphics[width=\textwidth]{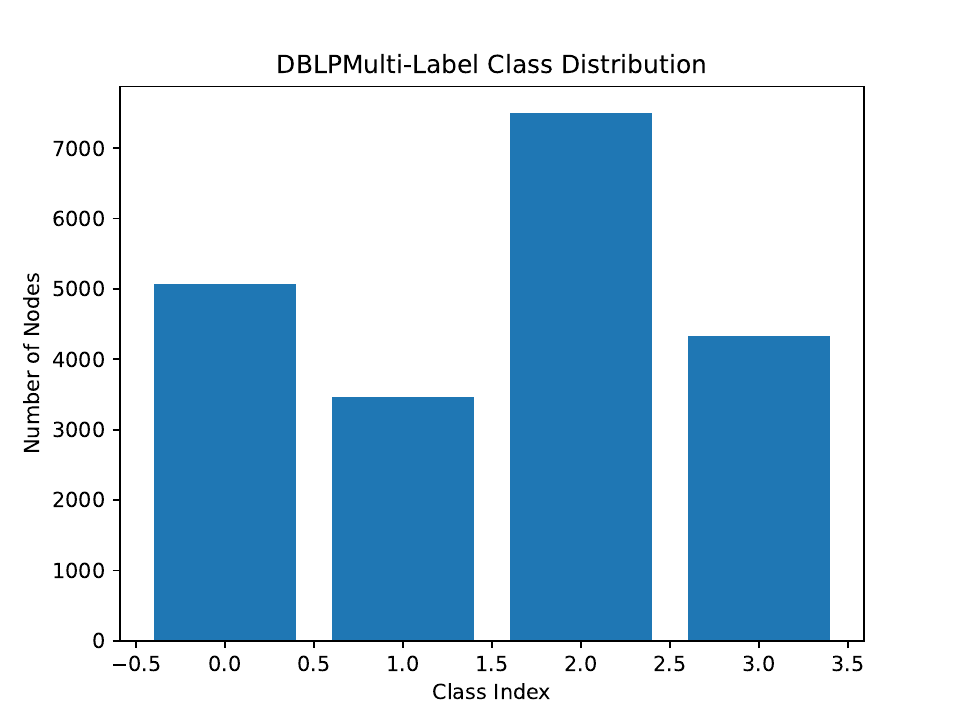}
        \caption{DBLP }
    \end{subfigure}
    \begin{subfigure}[b]{0.24\textwidth}
        \centering
        \includegraphics[width=\textwidth]{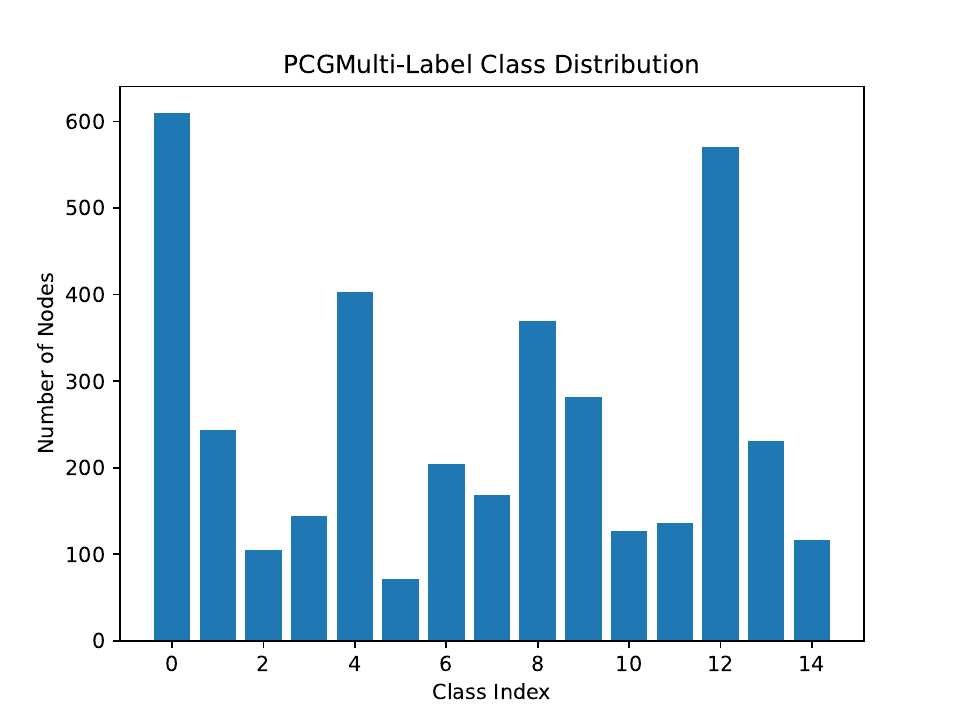}
        \caption{PCG }
    \end{subfigure}
    \begin{subfigure}[b]{0.24\textwidth}
        \centering
        \includegraphics[width=\textwidth]{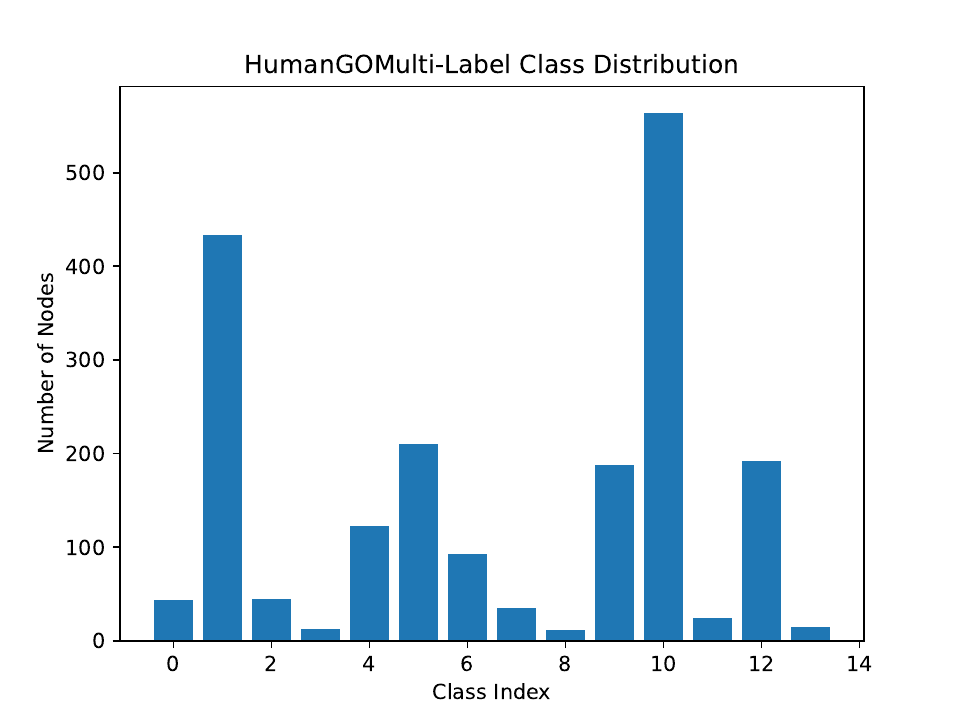}
        \caption{HumanGO }
    \end{subfigure}
    \begin{subfigure}[b]{0.24\textwidth}
        \centering
        \includegraphics[width=\textwidth]{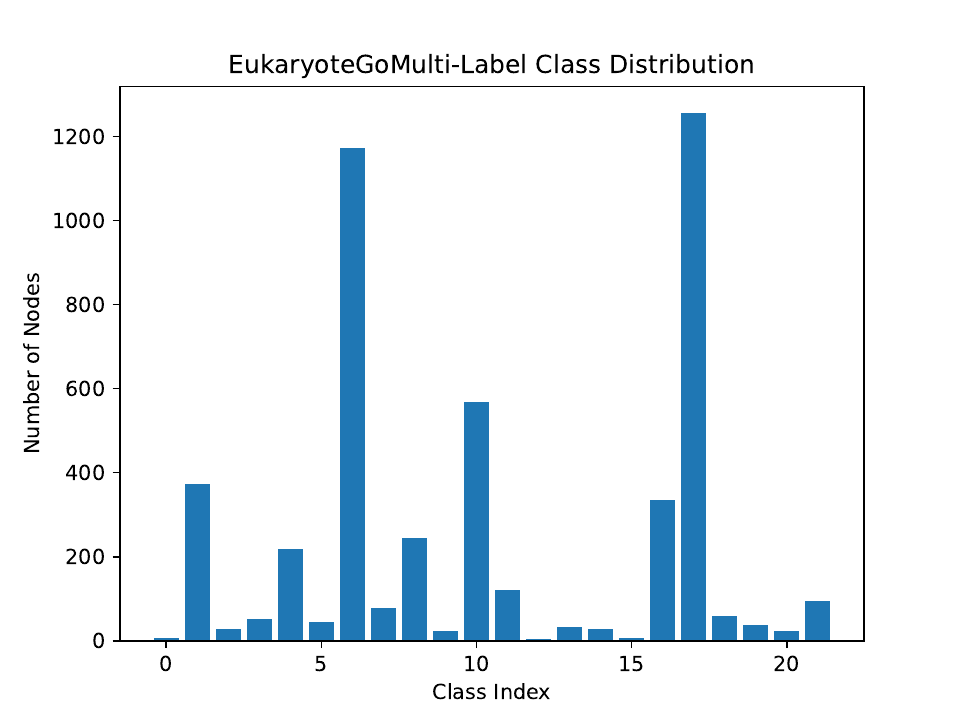}
        \caption{EukaryoteGo }
    \end{subfigure}
    
    \caption{Multi-label Class Distribution Visualization}
    \label{ap:class}
\end{figure*}

Following the above definition, we report the visualizations of different datasets as follows. The label correlation and class distribution are shown in Figures \ref{ap:label} and \ref{ap:class}, respectively. We find that the more complex the labels, the more the condensation methods would rely on structure and become more random. For future work, more suitable methods need to be fit to multi-label scenarios.

\subsection{F1-Macro Results of Adaptation}
\begin{figure}[ht]
    \centering
    \includegraphics[width=\textwidth]{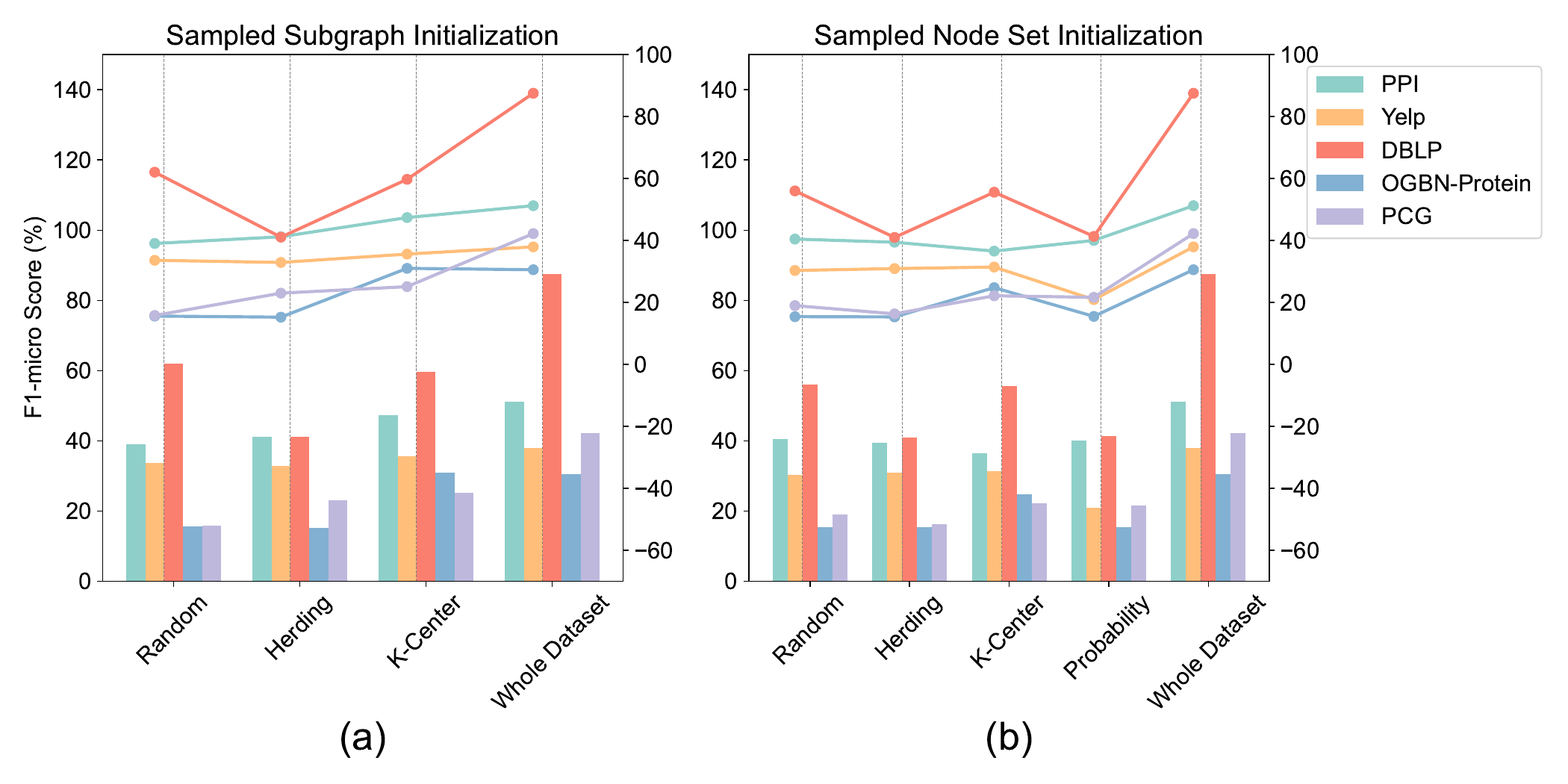}
    \caption{Different Initialization Methods Performance. Performance metrics of the model, with the F1-score represented as a decimal value. }
    \label{fig:coreset}
\end{figure}

\begin{figure}[ht]
    \centering
    \includegraphics[width=\textwidth]{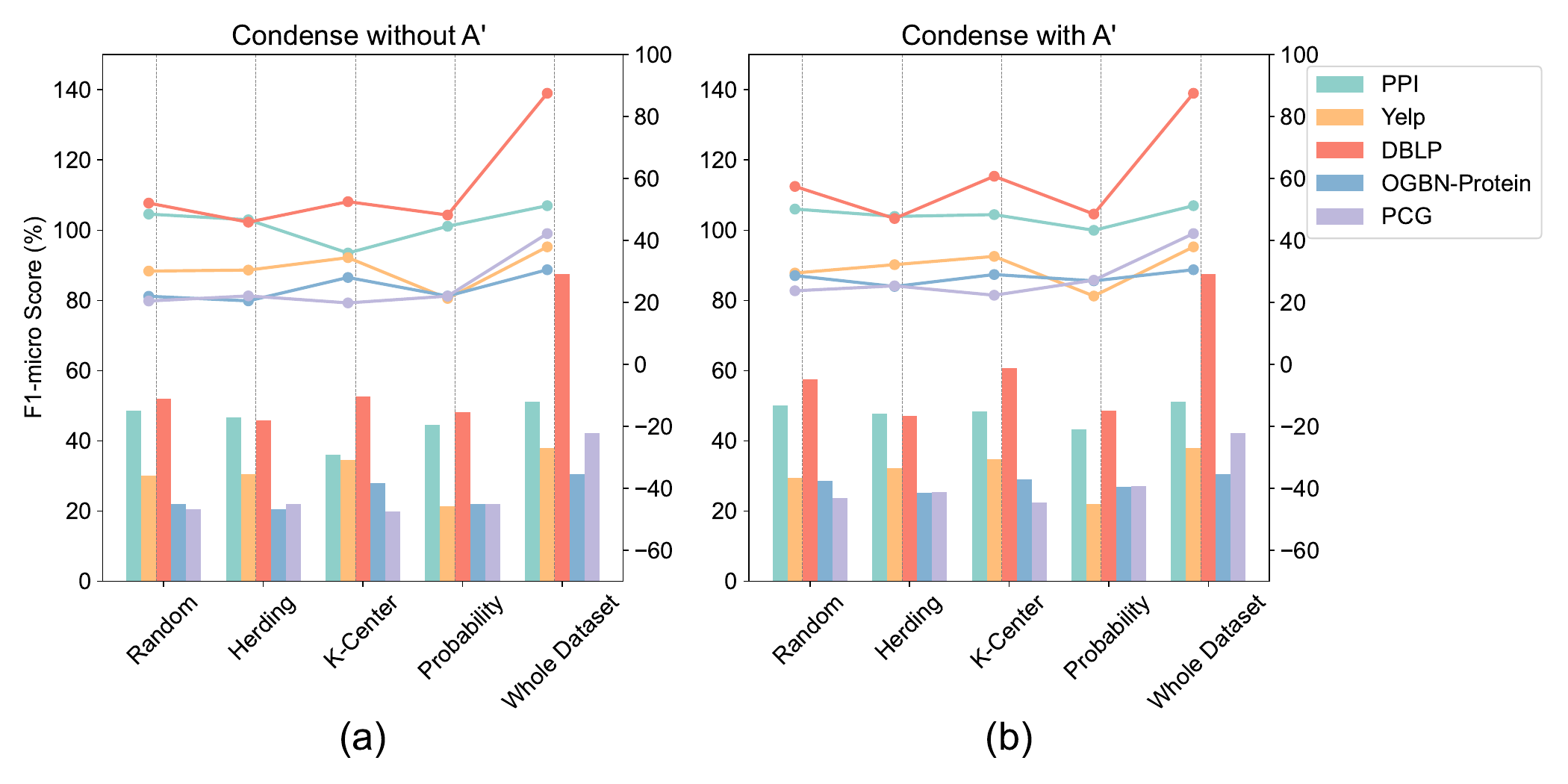}
    \caption{Performance Optimized by SoftMargin Loss. Performance metrics of the model, with the F1-score represented as a decimal value. }
    \label{fig:loss1}
\end{figure}

\begin{figure}[ht]
    \centering
    \includegraphics[width=\textwidth]{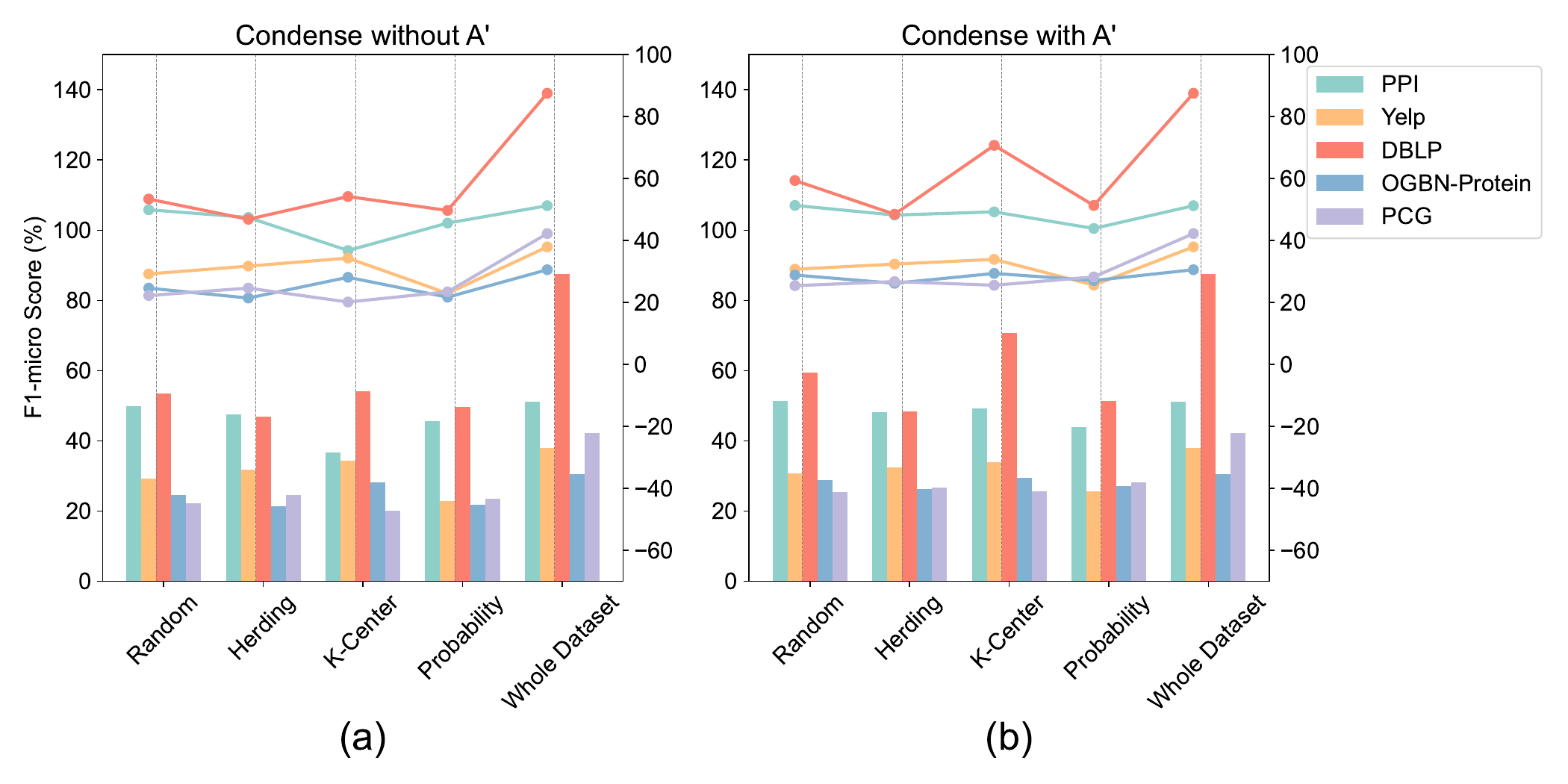}
    \caption{Performance Optimized by BCE Loss. Performance metrics of the model, with the F1-score represented as a decimal value. }
    \label{fig:BCE}
\end{figure}

Figures \ref{fig:coreset}, \ref{fig:loss1} and \ref{fig:BCE} show the visualizations of different adapation strategies measured by F1-micro score. To further investigate the effectiveness of the methods, we also report the F1-macro score in Tables \ref{ap:tab-init}
, \ref{ap:tab-loss1} and \ref{ap:tab-BCE}.

\begin{table*}[t]
    \footnotesize
    \centering
    \resizebox{\textwidth}{!}{
    \begin{tabular}{c c c c c c c c c c}
    \hline
    \multirow{2}{*}{\bf Datasets} & \multirow{2}{*}{\bf C-rate} & \multicolumn{2}{c}{\bf Random} & \multicolumn{2}{c}{\bf Herding} & \multicolumn{2}{c}{\bf K-Center} & \multirow{2}{*}{\bf Probability} & \multirow{2}{*}{\bf Whole Dataset}\\
    \cline{3-8}
    & & \bf Subgraph & \bf Nodes & \bf Subgraph & \bf Nodes & \bf Subgraph & \bf Nodes & & \\
     \hline
    \bf PPI & \bf 1.00\% & 12.58 & 10.67 & 11.66 & 17.98 & 17.36 &\bf 18.37 & 12.55 & 30.06 \\
    \bf Yelp & \bf 0.02\% & 5.09 & 3.76 & 5.64 & 3.66 &\bf 12.30 & 8.78 & 2.87 & 15.18 \\
    \bf DBLP & \bf 0.80\% & 56.00 & 51.43 & 15.63 & 15.41 &\bf 56.90 & 52.74 & 15.68 & 86.39 \\
    \bf OGBN-Pro & \bf 0.1\% & 2.12 & 7.12 & 5.21 & 7.17 &\bf 10.13 & 6.42 & 6.36 & 9.16 \\
    \bf PCG & \bf 4\% & 5.20 & 9.69 &\bf 13.84 & 6.45 & 5.73 & 7.10 & 11.35 & 31.49 \\
    \hline
    \end{tabular}
    }
    \footnotesize
    \caption{F1 Macro Score(\%) of Coreset Method with Different Initializetion Strategies}
    \label{ap:tab-init}
\end{table*}

\begin{table*}
    \centering
    \footnotesize
    \resizebox{\textwidth}{!}{
    \begin{tabular}{c c c c c c c c c c c}
    \hline
    \multirow{2}{*}{\bf Datasets} & \multirow{2}{*}{\bf C-rate} & \multicolumn{2}{c}{\bf Random} & \multicolumn{2}{c}{\bf Herding} & \multicolumn{2}{c}{\bf K-Center} & \multicolumn{2}{c}{\bf Probability} & \multirow{2}{*}{\bf Whole Dataset}\\
    \cline{3-10}
    & & \bf Without $A'$ & \bf With $A'$ & \bf Without $A'$ & \bf With $A'$ & \bf Without $A'$ &\bf With $A'$ &\bf Without $A'$ &\bf With $A'$ & \\
    \hline
    \textbf{PPI} & \textbf{1.00\%} & 22.60 & 28.33 & \bf 30.37 & 18.32 & 11.62 & 23.12 & 16.66 & 12.61 & 30.06 \\
    \textbf{Yelp} & \textbf{0.02\%} & 3.72 & 6.98 & 3.57 & 3.99 & 6.37 &\bf 9.02 & 2.13 & 1.86 & 15.18 \\
    \textbf{DBLP} & \textbf{1\%} & 35.81 & 37.84 & 27.08 & 27.11 & 32.21 &\bf 49.36 & 42.33 & 38.08 & 86.39 \\
    \textbf{OGBN-Pro} & \textbf{0.10\%} & 4.36 & 5.47 & 5.37 &\bf 7.18 & 5.26 & 5.32 & 3.21 & 4.76 & 9.16 \\
    \textbf{PCG} & \textbf{4\%} & 12.42 & 13.86 & 10.09 &\bf 15.50 & 7.34 & 9.47 & 13.29 & 6.50 & 31.49 \\
    \hline
    \end{tabular}
    }
    \captionsetup{justification=centering}
    \caption{F1-Macro Score (\%) of GCond Method with Random/Herding/K-Center/Probability Distribution Initialization with/without Learning from Graph Structure for SoftMarginLoss}
    \label{ap:tab-loss1}
    \centering
    \resizebox{\textwidth}{!}{
    \begin{tabular}{c c c c c c c c c c c}
    \hline
    \multirow{2}{*}{\bf Datasets} & \multirow{2}{*}{\bf C-rate} & \multicolumn{2}{c}{\bf Random} & \multicolumn{2}{c}{\bf Herding} & \multicolumn{2}{c}{\bf K-Center} & \multicolumn{2}{c}{\bf Probability} & \multirow{2}{*}{\bf Whole Dataset}\\
    \cline{3-10}
    & & \bf Without $A'$ & \bf With $A'$ & \bf Without $A'$ & \bf With $A'$ & \bf Without $A'$ &\bf With $A'$ &\bf Without $A'$ &\bf With $A'$ & \\
    \hline
    \textbf{PPI} & \textbf{1.00\%} & 22.82 &\bf 23.86 & 21.72 & 21.35 & 20.69 & 21.47 & 17.68 & 20.62 & 30.06 \\
    \textbf{Yelp} & \textbf{0.02\%} & 3.57 &\bf 8.02 & 4.24 & 4.13 & 6.13 & 6.22 & 4.00 & 3.07 & 15.18 \\
    \textbf{DBLP} & \textbf{1\%} & 34.89 & 51.44 & 26.94 & 28.28 & 34.87 &\bf 68.44 & 38.36 & 39.87 & 86.39 \\
    \textbf{OGBN-Pro} & \textbf{0.10\%} & 6.79 & 6.19 & 5.87 & 7.30 & 4.98 &\bf 7.47 & 2.89 & 4.99 & 9.16 \\
    
    \textbf{PCG} & \textbf{4\%} & 7.83 & 7.29 &\bf 15.17 & 14.27 & 8.02 & 13.32 & 13.13 & 11.30 & 31.49 \\
    \hline
    \end{tabular}
    }
    \caption{F1-Macro Score (\%) of GCond Method with Random/Herding/K-Center/Probability Distribution Initialization with/without Learning from Adjacent Nodes for BCELoss}
    \label{ap:tab-BCE}

\end{table*}

\subsection{Class-Weighted Optimization}

Let $N_{\max} = \max_{c\in {0,\cdots, C-1}}N_c$ is the maximum number of samples across all classes. For each class $c$ the class-wise coefficient is defined as:
\begin{equation}
    \alpha_c = \frac{N_c}{N_{\max}}
\end{equation}
The total loss for all the $N$ nodes would be the sum over all the classes and their respective nodes $N_c$:
\begin{equation}
    \mathcal{L}_{single-label} = \sum_{c=0}^{C-1} \alpha_c \sum_{j\in N_c}\ell_{CE}(z_j, y_j)
\end{equation}
This is the general equation for different methods, as the condensation part is using the $\mathcal{M}(\cdot)$ as the matching loss, for example, GCond would compute the gradient for each class with class-wise coefficient and sum them up to get the final loss. 

Furthermore, for each task we are working on the binary classification problem. To better capture the contribution of each class in multi-label, we introduce class weights $\omega \in \mathbb{R}^K$ for each task $k$:
\begin{equation}
    \begin{aligned}
        \ell_\text{BCE}(z_{j,k}, y_{j,k}) = -\omega_k(y_{j,k} \log(\sigma(z_{j,k})) + (1-y_{j,k})\log(\sigma(z_{j,k}))),
    \end{aligned}
\end{equation}
where $\omega_k$ is the positive class weight, scaling the loss for each positive class $y_{i, k} = 1$.
Therefore, the final loss of multi-label classification is:
\begin{equation}
    \begin{aligned}
        \mathcal{L}_{multi-label} = -\frac{1}{N}\sum_{j=0}^{N-1}\sum_{k=0}^{K-1}\omega_k(y_{j,k} \log(\sigma(z_{j,k})) \\
        + (1-y_{j,k})\log(1 - \sigma(z_{j,k})))
    \end{aligned}
\end{equation}

\begin{figure}[ht]
    \centering
    \includegraphics[width=\textwidth]{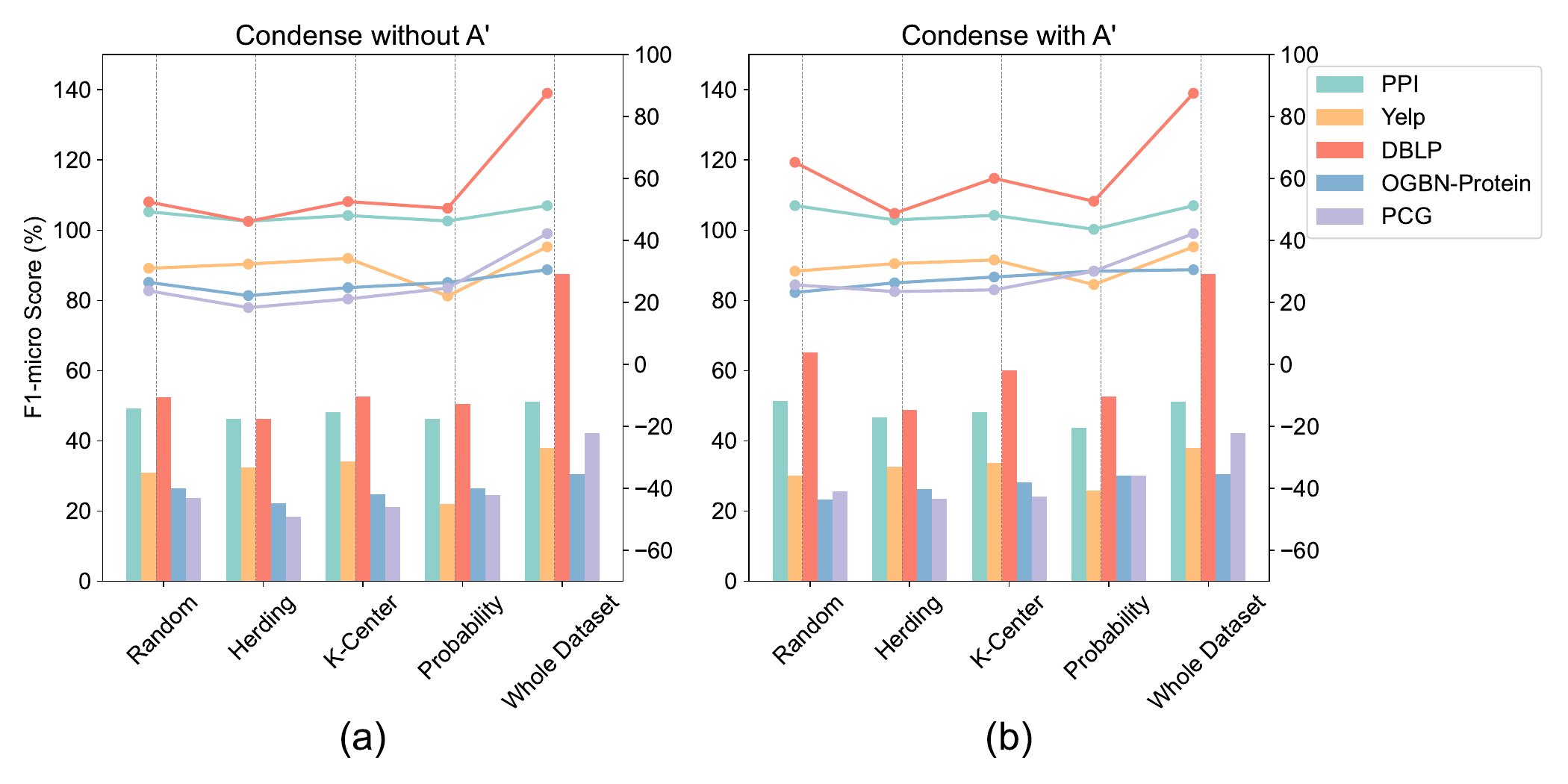}
    \caption{Performance Optimized by BCE Loss with Class Weight $\omega$}
    \label{fig:BCE+}
\end{figure}

This is helpful when there is an imbalance between the number of positive and negative samples. The F1-micro results are shown in Table \ref{ap:tab-BCE+0} and Figure \ref{fig:BCE+}. F1-macro results are shown in Table \ref{ap:tab-BCE+1}.

\begin{table*}    
    \centering
    \resizebox{\textwidth}{!}{
    \begin{tabular}{c c c c c c c c c c c}
    \hline
    \multirow{2}{*}{\bf Datasets} & \multirow{2}{*}{\bf C-rate} & \multicolumn{2}{c}{\bf Random} & \multicolumn{2}{c}{\bf Herding} & \multicolumn{2}{c}{\bf K-Center} & \multicolumn{2}{c}{\bf Probability} & \multirow{2}{*}{\bf Whole Dataset}\\
    \cline{3-10}
    & & \bf Without $A'$ & \bf With $A'$ & \bf Without $A'$ & \bf With $A'$ & \bf Without $A'$ &\bf With $A'$ &\bf Without $A'$ &\bf With $A'$ & \\
    \hline
    \textbf{PPI} & \textbf{1.00\%} & 49.30 &\bf 51.28 & 46.25 & 46.65 & 48.10 & 48.15 & 46.32 & 43.64 & 51.26 \\
    \textbf{Yelp} & \textbf{0.02\%} & 31.06 & 30.12 & 32.42 & 32.58 &\bf 34.26 & 33.75 & 22.10 & 25.82 & 37.97 \\
    \textbf{DBLP} & \textbf{0.80\%} & 52.47 &\bf 65.27 & 46.19 & 48.79 & 52.58 & 60.09 & 50.43 & 52.69 & 87.55 \\
    \textbf{OGBN-Proteins} & \textbf{0.10\%} & 26.51 & 23.25 & 22.25 & 26.36 & 24.81 & 28.24 & 26.47 &\bf 30.13 & 30.59 \\
    \textbf{PCG} & \textbf{4\%} & 23.78 & 25.72 & 18.36 & 23.53 & 21.15 & 24.10 & 24.68 &\bf 30.13 & 42.26 \\
    \hline
    \end{tabular}
    }
    \caption{F1-Micro Score (\%) of GCond Method with Random/Herding/K-Center/Probability Distribution Initialization with/without Learning from Adjacent Nodes for BCE Loss and Balanced Coefficient}
\label{ap:tab-BCE+0}
\end{table*}

\begin{table}
    \centering
    \resizebox{\textwidth}{!}{
    \begin{tabular}{c c c c c c c c c c c}
    \hline
    \multirow{2}{*}{\bf Datasets} & \multirow{2}{*}{\bf C-rate} & \multicolumn{2}{c}{\bf Random} & \multicolumn{2}{c}{\bf Herding} & \multicolumn{2}{c}{\bf K-Center} & \multicolumn{2}{c}{\bf Probability} & \multirow{2}{*}{\bf Whole Dataset}\\
    \cline{3-10}
    & & \bf Without $A'$ & \bf With $A'$ & \bf Without $A'$ & \bf With $A'$ & \bf Without $A'$ &\bf With $A'$ &\bf Without $A'$ &\bf With $A'$ & \\
    \hline
    \textbf{PPI} & \textbf{1.00\%} & 21.75 & 25.65 & 15.55 & 27.21 & 27.09 &\bf 31.97 & 14.31 & 21.43 & 30.06 \\
    \textbf{Yelp} & \textbf{0.02\%} & 3.91 & 6.49 & 4.06 & 5.16 &\bf 11.72 & 7.78 & 1.86 & 2.42 & 15.18 \\
    \textbf{DBLP} & \textbf{0.80\%} & 32.79 &\bf 61.83 & 26.44 & 29.25 & 32.64 & 56.29 & 37.78 & 42.11 & 86.39 \\
    \textbf{OGBN-Pro} & \textbf{0.10\%} & 4.60 & 6.05 & 3.54 & 4.69 & 6.42 & 7.89 & 4.38 &\bf 8.56 & 9.16 \\
    \textbf{PCG} & \textbf{4\%} & 13.26 & 11.93 & 12.24 &\bf 15.75 & 5.70 & 9.88 & 11.60 & 8.56 & 31.49 \\
    \hline
    \end{tabular}
    }
    \caption{F1-Macro Score (\%) of GCond Method with Random/Herding/K-Center/Probability Distribution Initialization with/without Learning from Adjacent Nodes for BCE Loss and Balanced Coefficient}
    \label{ap:tab-BCE+1}
\end{table}

\end{document}